# Narrowing the Modeling Gap:
# A Cluster-Ranking Approach to Coreference Resolution

**Altaf Rahman**                                                    ALTAF@HLT.UTDALLAS.EDU
**Vincent Ng**                                                      VINCE@HLT.UTDALLAS.EDU
*Human Language Technology Research Institute*
*University of Texas at Dallas*
*800 West Campbell Road; Mail Station EC31*
*Richardson, TX 75080-3021 U.S.A.*

## Abstract

Traditional learning-based coreference resolvers operate by training the *mention-pair* model for determining whether two mentions are coreferent or not. Though conceptually simple and easy to understand, the mention-pair model is linguistically rather unappealing and lags far behind the heuristic-based coreference models proposed in the pre-statistical NLP era in terms of sophistication. Two independent lines of recent research have attempted to improve the mention-pair model, one by acquiring the *mention-ranking* model to rank preceding mentions for a given anaphor, and the other by training the *entity-mention* model to determine whether a preceding cluster is coreferent with a given mention. We propose a cluster-ranking approach to coreference resolution, which combines the strengths of the mention-ranking model and the entity-mention model, and is therefore theoretically more appealing than both of these models. In addition, we seek to improve cluster rankers via two extensions: (1) lexicalization and (2) incorporating knowledge of anaphoricity by jointly modeling anaphoricity determination and coreference resolution. Experimental results on the ACE data sets demonstrate the superior performance of cluster rankers to competing approaches as well as the effectiveness of our two extensions.

## 1. Introduction

Noun phrase (NP) coreference resolution is the task of identifying which NPs (or *mentions* in ACE terminology[1]) in a text or dialogue refer to the same real-world entity or concept. From a computational perspective, coreference is a *clustering* task, with the goal of partitioning a set of mentions into coreference clusters where each cluster contains all and only the mentions that are co-referring. From a mathematical perspective, a coreference relation is an *equivalence* relation defined on a pair of mentions, as it satisfies reflexivity, symmetry, and transitivity. Following our previous work on coreference resolution, we use the term *anaphoric* to describe a mention that is part of a coreference chain but is not the head of a chain. Given an anaphoric mention $m_k$, an *antecedent* of $m_k$ is a mention that is coreferent with $m_k$ and precedes it in the associated text, and the set of *candidate antecedents* of $m_k$ consists of all mentions that precede $m_k$.[2]

---

1. More precisely, a *mention* is an instance of reference to an *entity* in the real world. In this article, we treat the terms *mention* and *noun phrase* as synonymous and use them interchangeably.
2. Note that these definitions are somewhat overloaded. Linguistically, an anaphor is a noun phrase that depends on its antecedent for its semantic interpretation. Hence, "Barack Obama" can be anaphoric in our definition but not in the formal definition.

The research focus of computational coreference resolution exhibited a gradual shift from heuristic-based approaches to machine learning approaches in the past decade. The shift can be attributed in part to the advent of the statistical natural language processing (NLP) era, and in part to the public availability of coreference-annotated corpora produced as a result of the MUC-6 and MUC-7 conferences and the series of ACE evaluations. One of the most influential machine learning approaches to coreference resolution is the classification-based approach, where coreference is recast as a binary classification task (e.g., Aone & Bennett, 1995; McCarthy & Lehnert, 1995). Specifically, a classifier that is trained on coreference-annotated data is used to determine whether a pair of mentions is co-referring or not. However, the pairwise classifications produced by this classifier (which is now commonly known as the *mention-pair* model) may not satisfy the transitivity property inherent in the coreference relation, since it is possible for the model to classify (A,B) as coreferent, (B,C) as coreferent, and (A,C) as not coreferent. As a result, a separate clustering mechanism is needed to coordinate the possibly contradictory pairwise classification decisions and construct a partition of the given mentions.

The mention-pair model has significantly influenced learning-based coreference research in the past fifteen years. In fact, many of the recently published coreference papers are still based on this classical learning-based coreference model (e.g., Bengtson & Roth, 2008; Stoyanov, Gilbert, Cardie, & Riloff, 2009). Despite its popularity, the model has at least two major weaknesses. First, since each candidate antecedent for a mention to be resolved (henceforth an *active mention*) is considered independently of the others, this model only determines how good a candidate antecedent is relative to the active mention, but not how good a candidate antecedent is relative to other candidates. In other words, it fails to answer the critical question of which candidate antecedent is most probable. Second, it has limitations in its expressiveness: the information extracted from the two mentions alone may not be sufficient for making an informed coreference decision, especially if the candidate antecedent is a pronoun (which is semantically empty) or a mention that lacks descriptive information such as gender (e.g., "Clinton").

Recently, coreference researchers have investigated alternative models of coreference that aim to address the aforementioned weaknesses of the mention-pair model. To address the first weakness, researchers have proposed the *mention-ranking* model. This model determines which candidate antecedent is most probable given an active mention by imposing a ranking on its candidate antecedents (e.g., Denis & Baldridge, 2007b, 2008; Iida, Inui, & Matsumoto, 2009). Ranking is arguably a more natural formulation of coreference resolution than classification, as a ranker allows all candidate antecedents to be considered *simultaneously* and therefore directly captures the competition among them. Another desirable consequence is that there exists a natural resolution strategy for a ranking approach: a mention is resolved to the candidate antecedent that has the highest rank. This contrasts with classification-based approaches, where many clustering algorithms have been employed to co-ordinate the pairwise coreference decisions (because it is unclear which one is the best). To address the second weakness, researchers have proposed the *entity-mention* coreference model (e.g., Luo, Ittycheriah, Jing, Kambhatla, & Roukos, 2004; Yang, Su, Zhou, & Tan, 2004; Yang, Su, Lang, Tan, & Li, 2008). Unlike the mention-pair model, the entity-mention model is trained to determine whether an active mention belongs to a preceding, possibly partially-formed, coreference cluster. Hence, it can employ *cluster-level* features (i.e., fea-tures that are defined over any subset of mentions in a preceding cluster), which makes it more expressive than the mention-pair model.

While the entity-mention model and the mention-ranking model are conceptually simple extensions to the mention-pair model, they were born nearly ten years after the mention-pair model was proposed, and in particular, their contributions should not be under-estimated: they paved a new way of thinking about supervised modeling of coreference that represents a significant departure from their mention-pair counterpart, which for many years is *the* learning-based coreference model for NLP researchers. The proposal of these two models is facilitated in part by advances in statistical modeling of natural languages: statistical NLP models have evolved from capturing local information to global information, and from employing classification-based models to ranking-based models. In the context of coreference resolution, the entity-mention model enables us to compute features based on a *variable* number of mentions, and the mention-ranking model enables us to *rank* a *variable* number of candidate antecedents. Nevertheless, neither of these models addresses both weaknesses of the mention-pair model satisfactorily: while the mention-ranking model allows all candidate antecedents to be ranked and compared simultaneously, it does not enable the use of cluster-level features; on the other hand, while the entity-mention model can employ cluster-level features, it does not allow all candidates to be considered simultaneously.

Motivated in part by this observation, we propose a learning-based approach to coreference resolution that is theoretically more appealing than both the mention-ranking model and the entity-mention model: the *cluster-ranking* approach. Specifically, we recast coreference as the problem of determining which of a set of preceding coreference *clusters* is the best to link to an active mention using a learned *cluster-ranking model*. In essence, the cluster-ranking model combines the strengths of the mention-ranking model and the entity-mention model, and addresses *both* weaknesses associated with the mention-pair model.

While the cluster-ranking model appears to be a conceptually simple and natural extension of the entity-mention model and the mention-ranking model, we believe that such simplicity stems primarily from our choice of a presentation of these concepts that is easiest for the reader to understand. In particular, we note that the mental processes involved in the design of the cluster-ranking model are by no means as simple as the way the model is presented: it requires not only an analysis of the strengths and weaknesses of existing approaches to learning-based coreference resolution and the connection between them, but also our formulation of the view that the entity-mention model and the mention-ranking model are addressing two complementary weaknesses of the mention-pair model. We believe that the significance of our cluster-ranking model lies in bridging two rather independent lines of learning-based coreference research that have been going on in the past few years, one involving the entity-mention model and the other the mention-ranking model.

In addition, we seek to improve the cluster-ranking model with two sources of linguistic knowledge. First, we propose to exploit knowledge of *anaphoricity* (i.e., knowledge of whether a mention is anaphoric or not). Anaphoricity determination is by no means a new problem, and neither is the use of anaphoricity information to improve coreference resolution. Our innovation lies in the way we *learn* knowledge of anaphoricity. Specifically, while previous work has typically adopted a *pipeline* coreference architecture, in which anaphoricity determination is performed prior to coreference resolution and the resulting information is used to prevent a coreference system from resolving mentions that are de-termined to be non-anaphoric (for an overview, see the work of Poesio, Uryupina, Vieira, Alexandrov-Kabadjov, & Goulart, 2004), we propose a model for *jointly* learning anaphoricity determination and coreference resolution. Note that a major weakness of the pipeline architecture lies in the fact that errors in anaphoricity determination could be propagated to the coreference resolver, possibly leading to a deterioration of coreference performance (Ng & Cardie, 2002a). Our joint model is a potential solution to this error-propagation problem.

Second, we examine a kind of linguistic features that is not exploited by the majority of existing supervised coreference resolvers: *word pairs* that are composed of the strings (or the head nouns) of an active mention and one of its preceding mentions. Intuitively, these word pairs contain useful information. For example, they may help improve the *precision* of a model, by allowing a learner to learn that "it" only has a moderate probability of being anaphoric, and that "the contrary" taken from the phrase "on the contrary" is never anaphoric. They may also help improve its recall, by allowing the learner to determine, for instance, that "airline" and "carrier" can be coreferent. Hence, they offer a convenient means to attack one of the major problems in coreference research: identifying coreferent common nouns that are lexically dissimilar but semantically related. Note that they are extremely easy to compute, even more so than the so-called "cheap" features such as string-matching and grammatical features (Yang, Zhou, Su, & Tan, 2003), but the majority of the existing supervised coreference systems are *unlexicalized* and hence are not exploiting them. Somewhat unexpectedly, however, for researchers who do lexicalize their coreference models by employing word pairs as features (e.g., Luo et al., 2004; Daumé III & Marcu, 2005; Bengtson & Roth, 2008), their feature analysis experiments indicate that lexical features are at best marginally useful. For instance, Luo et al. and Daume III and Marcu report that leaving out lexical features in their feature ablation experiments causes the ACE value to drop only by 0.8 and 0.7, respectively. While previous attempts on lexicalization merely append all word pairs to a conventional coreference feature set, our goal is to investigate whether we can make better use of lexical features for learning-based coreference resolution.

To sum up, we propose a cluster-ranking approach to coreference resolution and a joint model for exploiting anaphoricity information, and investigate the role of lexicalization in learning-based coreference resolution. Besides empirically demonstrating that our cluster-ranking model significantly outperforms competing approaches on the ACE 2005 coreference data set, and that our two extensions to the model, namely lexicalization and joint modeling, are effective in improving its performance, we believe our work makes four contributions to coreference resolution:

**Narrowing the modeling gap.** While machine learning approaches to coreference resolution have received a lot of attention since the mid-1990s, the mention-pair model has heavily influenced learning-based coreference research for more than a decade, and yet this model lags far behind the heuristic-based coreference models proposed in the 1980s and 1990s in terms of sophistication. In particular, the notion of ranking can be traced back to centering algorithms (for more information, see the books by Mitkov, 2002; Walker, Joshi, & Prince, 1998), and the idea behind ranking preceding clusters (in a heuristic manner) can be found in Lappin and Leass's (1994) influential paper on pronoun resolution. While our cluster-ranking model does not completely close the gap between the simplicity of machine learning approaches and the sophistication of heuristic approaches to coreference resolu-tion, we believe that it represents an important step towards narrowing this gap. Another important gap that our cluster-ranking model helps to bridge is the two independent lines of learning-based coreference research that have been going on in the past few years, one involving the entity-mention model and the other mention-ranking model.

**Promoting the use of ranking models.** While the mention-ranking model has been empirically shown to outperform the mention-pair model (Denis & Baldridge, 2007b, 2008), the former has not received as much attention among coreference researchers as it should. In particular, the mention-pair model continues to be more popularly used and investigated in the past few years than the mention-ranking model. We believe the lack of excitement for ranking-based approaches to coreference resolution can be attributed at least in part to a lack of theoretical understanding of ranking, as previous work on ranking-based coreference resolution has employed ranking algorithms essentially as a black box. Without opening the black box, it could be difficult for researchers to appreciate the subtle difference between ranking and classification. In an attempt to promote the use of ranking-based models, we provide a brief history of the use of ranking in coreference resolution (Section 2), and tease apart the differences between classification and ranking by showing the constrained optimization problem a support vector machine (SVM) attempts to solve in classification-based and ranking-based coreference models (Section 3).

**Gaining a better understanding of existing learning-based coreference models.** Recall that lexicalization is one of the two linguistic knowledge sources that we propose to use to improve the cluster-ranking model. Note that lexicalization can be applied to not only the cluster-ranking model, but essentially any learning-based coreference models. However, as mentioned before, the vast majority of existing coreference resolvers are unlexicalized. In fact, the mention-ranking model has only been shown to improve the mention-pair model on an unlexicalized feature set. In an attempt to gain additional insights into the behavior of different learning-based coreference models, we compare their performance on a lexicalized feature set. Furthermore, we analyze them via experiments involving feature ablation and data source adaptability, as well as report their performance on resolving different types of anaphoric expressions.

**Providing an implementation of the cluster-ranking model.** To stimulate further research on ranking-based approaches to coreference resolution, and to facilitate the use of coreference information in high-level NLP applications, we make our software that implements the cluster-ranking model publicly available.[3]

The rest of this article is organized as follows. Section 2 provides an overview of the use of ranking in coreference resolution. Section 3 describes our baseline coreference models: the mention-pair model, the entity-mention model, and the mention-ranking model. We discuss our cluster-ranking approach and our joint model for anaphoricity determination and coreference resolution in Section 4. Section 5 provides the details of how we lexicalize the coreference models. We present evaluation results and experimental analyses of different aspects of the coreference models in Section 6 and Section 7, respectively. Finally, we conclude in Section 8.

---

3. The software is available at `http://www.hlt.utdallas.edu/~altaf/cherrypicker/`.

## 2. Ranking Approaches to Coreference Resolution: A Bit of History

While ranking is theoretically and empirically a better formulation of learning-based coreference resolution than classification, the mention-ranking model has not been as popularly used and investigated as its mention-pair counterpart since it was proposed. To promote ranking-based coreference models, and to set the stage for further discussion of learning-based coreference models in the next section, we provide in this section a brief history of the use of ranking in heuristic-based and learning-based coreference resolution.

In a broader sense, many heuristic anaphora and coreference resolvers are ranking-based. For example, to find an antecedent for an anaphoric pronoun, Hobbs's (1978) seminal syntax-based resolution algorithm considers the sentences in a given text in reverse order, starting from the sentence in which the pronoun resides and searching for potential antecedents in the corresponding parse trees in a left-to-right, breadth-first manner that obeys binding and agreement constraints. Hence, if we keep searching until the beginning of the text is reached (i.e., we do not stop even after the algorithm proposes an antecedent), we will obtain a ranking of the candidate antecedents for the pronoun under consideration, where the rank of a candidate is determined by the order in which it is proposed by the algorithm. In fact, the rank of an antecedent obtained via this method is commonly known as its Hobbs's distance, which has been used as a linguistic feature in statistical pronoun resolvers (e.g., Ge, Hale, & Charniak, 1998; Charniak & Elsner, 2009). In general, search-based resolution algorithms like Hobbs's consider candidate antecedents in a particular order and (typically) propose the first candidate that satisfies all linguistic constraints as the antecedent.

Strictly speaking, however, we may want to consider a heuristic resolution algorithm as a ranking-based algorithm only if it considers all candidate antecedents simultaneously, for example by assigning a rank or score to each candidate and selecting the highest-ranked or highest-scored candidate to be the antecedent. Even under this stricter definition of ranking, there are still many heuristic resolvers that are ranking-based. These resolvers typically assign a rank or score to each candidate antecedent based on a number of *factors*, or knowledge sources, and then propose the one that has the highest rank or score as an antecedent (e.g., Carbonell & Brown, 1988; Cardie & Wagstaff, 1999). A factor belongs to one of two types: *constraints* and *preferences* (Mitkov, 1998). Constraints must be satisfied before two mentions can be posited as coreferent. Examples of constraints include gender and number agreement, binding constraints, and semantic compatibility. Preferences indicate the likelihood that a candidate is an antecedent. Some preference factors measure the compatibility between an anaphor and its candidate (e.g., *syntactic parallelism* favors candidates that have the same grammatical role as the anaphor), while other preference factors are computed based on the candidate only, typically capturing the salience of a candidate. Each constraint and preference is manually assigned a weight indicating its importance. For instance, gender disagreement is typically assigned a weight of $-\infty$, indicating that a candidate and the anaphor must agree in gender, whereas preference factors typically have a finite weight. The score of a candidate can then be obtained by summing the weights of the factors associated with the candidate.

Some ranking-based resolution algorithms do not assign a score to each candidate antecedent. Rather, they simply impose a ranking on the candidates based on their salience.

Perhaps the most representative family of algorithms that employ salience to rank candidates is centering algorithms (for descriptions of specific centering algorithms, see the work of Grosz, Joshi, & Weinstein, 1983, 1995; Walker et al., 1998; Mitkov, 2002), where the salience of a mention, typically estimated using its grammatical role, is used to rank forward-looking centers.

The work most related to ours is that of Lappin and Leass (1994), whose goal is to perform pronoun resolution by assigning an anaphoric pronoun to the highest-ranked preceding cluster, and is therefore a heuristic cluster-ranking model. Like many other heuristic-based resolvers, Lappin and Leass's algorithm identifies the highest-ranked preceding cluster for an active mention by first applying a set of linguistic constraints to filter candidate antecedents that are grammatically incompatible with the active mention, and then ranking the preceding clusters, which contain the mentions that survive the filtering process, using *salience factors*. Examples of salience factors include *sentence recency* (whether the preceding cluster contains a mention that appears in the sentence currently being processed), *subject emphasis* (whether the cluster contains a mention in the subject position), *existential emphasis* (whether the cluster contains a mention that is a predicate nominal in an existential construction), and *accusative emphasis* (whether the cluster contains a mention that appears in a verbal complement in accusative case). Each salience factor is associated with a manually-assigned weight that indicates its importance relative to other factors, and the score of a cluster is the sum of the weights of the salience factors that are applicable to the cluster. While Lappin and Leass's paper is a widely read paper on pronoun resolution, the cluster ranking aspect of their algorithm has rarely been emphasized. In fact, we are not aware of any recent work on learning-based coreference resolution that establishes the connection between the entity-mention model and Lappin and Leass's algorithm.

Despite the conceptual similarities, our cluster-ranking model and Lappin and Leass's (1994) algorithm differ in several respects. First, Lappin and Leass only tackle pronoun resolution rather than the full coreference task. Second, while they apply linguistic constraints to filter incompatible candidate antecedents, our resolution strategy is learned without applying hand-coded constraints in a separate filtering step. Third, while they attempt to compute the salience of a preceding cluster with respect to an active mention, we attempt to determine the *compatibility* between a cluster and an active mention, using factors that determine not only salience but also lexical and grammatical compatibility, for instance. Finally, their algorithm is heuristic-based, where the weights associated with each salience factor are encoded manually rather than learned, unlike our system.

The first paper on learning-based coreference resolution was written by Connolly, Burger, and Day (1994) and was published in the same year as Lappin and Leass's (1994) paper. Contrary to common expectation, the coreference model this paper proposes is a ranking-based model, not the influential mention-pair model. The main idea behind Connolly et al.'s approach is to convert a problem of ranking $N$ candidate antecedents into a set of *pairwise ranking* problems, each of which involves ranking exactly two candidates. To rank two candidates, a *classifier* can be trained using a training set where each instance corresponds to the active mention as well as two candidate antecedents and possesses a class value that indicates which of the two candidates is better. This idea is certainly ahead of its time, as it is embodied in many of the advanced ranking algorithms developed in the machine learning and information retrieval communities in the past few years. It is later re-invented at almost the same time, but independently, by Yang et al. (2003) and Iida, Inui, Takamura, and Matsumoto (2003), who refer to it as the *twin-candidate* model and the *tournament* model, respectively. The name *twin-candidate model* is motivated by the fact that the model considers two candidates at a time, whereas the name *tournament model* was assigned because each ranking of two candidates can be viewed as a tournament (with the higher-ranked candidate winning the tournament) and the candidate that wins the largest number of tournaments is chosen as the antecedent for the active mention. This bit of history is rarely mentioned in the literature, but it reveals three somewhat interesting and perhaps surprising facts. First, ranking was first applied to train coreference models much earlier than people typically think. Second, despite being the first learning-based coreference model, Connolly et al.'s ranking-based model is theoretically more appealing than the classification-based mention-pair model, and is later shown by Yang et al. and Iida et al.. to be empirically better as well. Finally, despite its theoretical and empirical superiority, Connolly et al.'s model was largely ignored by the NLP community and received attention only when it was re-invented nearly a decade later, while during this time period its mention-pair counterpart essentially dominated learning-based coreference research.[4]

We conclude this section by making the important observation that the distinction between classification and ranking applies to discriminative models but not generative models. Generative models try to capture the true conditional probability of some event. In the context of coreference resolution, this will be the probability of a mention having a particular antecedent or of it referring to a particular entity (i.e., preceding cluster). Since these probabilities have to normalize, this is similar to a ranking objective: the system is trying to raise the probability that a mention refers to the correct antecedent or entity at the expense of the probabilities that it refers to any other. Thus, the antecedent version of the generative coreference model as proposed by Ge et al. (1998) resembles the mention-ranking model, while the entity version as proposed by Haghighi and Klein (2010) is similar in spirit to the cluster-ranking model.

## 3. Baseline Coreference Models

In this section, we describe three coreference models that will serve as our baselines: the mention-pair model, the entity-mention model, and the mention-ranking model. For illustrative purposes, we will use the text segment shown in Figure 1. Each mention $m$ in the segment is annotated as $[m]^{cid}_{mid}$, where $mid$ is the mention id and $cid$ is the id of the cluster to which $m$ belongs. As we can see, the mentions are partitioned into four sets, with *Barack Obama*, *his*, and *he* in one cluster, and each of the remaining mentions in its own cluster.

---

4. It may not be possible (and perhaps not crucial) to determine why the mention-pair model received a lot more attention than Connolly et al.'s model, but since those were the days when academic papers could not be accessed easily in electronic form, we speculate that the publication venue played a role: Connolly et al.'s work was published in the New Methods in Language Processing conference in 1994 (and later as a book chapter in 1997), whereas the mention-pair model was introduced in Aone and Bennett's (1995) paper and McCarthy and Lehnert's (1995) paper, which appeared in the proceedings of two comparatively higher-profile AI conferences: ACL 1995 and IJCAI 1995.

---

[*Barack Obama*]$_1^1$ nominated [*Hillary Rodham Clinton*]$_2^2$ as [[*his*]$_3^1$ *secretary of state*]$_4^3$ on [*Monday*]$_5^4$. [*He*]$_6^1$ ...

---

Figure 1: An illustrative example

### 3.1 Mention-Pair Model

As noted before, the mention-pair model is a classifier that decides whether or not an active mention $m_k$ is coreferent with a candidate antecedent $m_j$. Each instance $\mathbf{i}(m_j, m_k)$ represents $m_j$ and $m_k$. In our implementation, an instance consists of the 39 features shown in Table 1. These features have largely been employed by state-of-the-art learning-based coreference systems (e.g., Soon, Ng, & Lim, 2001; Ng & Cardie, 2002b; Bengtson & Roth, 2008), and are computed automatically. As can be seen, the features are divided into four blocks. The first two blocks consist of features that describe the properties of $m_j$ and $m_k$, respectively, and the last two blocks of features describe the relationship between $m_j$ and $m_k$. The classification associated with a training instance is either positive or negative, depending on whether $m_j$ and $m_k$ are coreferent.

If one training instance were created from each pair of mentions, the negative instances would significantly outnumber the positives, yielding a skewed class distribution that will typically have an adverse effect on model training. As a result, only a subset of mention pairs will be generated for training. Following Soon et al. (2001), we create (1) a positive instance for each anaphoric mention $m_k$ and its closest antecedent $m_j$; and (2) a negative instance for $m_k$ paired with each of the intervening mentions, $m_{j+1}, m_{j+2}, \ldots, m_{k-1}$. In our running example shown in Figure 1, three training instances will be generated for *He*: $\mathbf{i}(Monday, He)$, $\mathbf{i}(secretary\ of\ state, He)$, and $\mathbf{i}(his, He)$. The first two of these instances will be labeled as negative, and the last one will be labeled as positive. To train the mention-pair model, we use the SVM learning algorithm from the SVM$^{light}$ package (Joachims, 1999).[5]

As mentioned in the introduction, while previous work on learning-based coreference resolution typically treats the underlying machine learner simply as a black-box tool, we choose to provide the reader with an overview of SVMs, the learner we are employing in our work. Note that this is a self-contained overview, but it is by no means a comprehensive introduction to maximum-margin learning: our goal here is to provide the reader with only the details that we believe are needed to understand the difference between classification and ranking and perhaps appreciate the importance of ranking.[6]

To begin with, assume that we are given a data set consisting of positively labeled points, which have a class value of +1, and negatively labeled points, which have a class

---

5. Since SVM$^{light}$ assumes real-valued features, it cannot operate on features with multiple discrete values directly. Hence, we need to convert the features shown in Table 1 into an equivalent set of features that can be used directly by SVM$^{light}$. For uniformity, we perform the conversion for *each* feature in Table 1 (rather than just the multi-valued features) as follows: we create one binary-valued feature for SVM$^{light}$ from each feature-value pair that can be derived from the feature set in Table 1. For example, PRONOUN_1 has two values, Y and N. So we will derive two binary-valued features, PRONOUN_1=Y and PRONOUN_1=N. One of them will have a value of 1 and the other will have a value of 0 for each instance.

6. For an overview of the theory of maximum-margin learning, we refer the reader to Burges's (1998) tutorial.

| **Features describing $m_j$, a candidate antecedent** | |
|---|---|
| 1 | PRONOUN_1 | Y if $m_j$ is a pronoun; else N |
| 2 | SUBJECT_1 | Y if $m_j$ is a subject; else N |
| 3 | NESTED_1 | Y if $m_j$ is a nested NP; else N |

| **Features describing $m_k$, the mention to be resolved** | |
|---|---|
| 4 | NUMBER_2 | SINGULAR or PLURAL, determined using a lexicon |
| 5 | GENDER_2 | MALE, FEMALE, NEUTER, or UNKNOWN, determined using a list of common first names |
| 6 | PRONOUN_2 | Y if $m_k$ is a pronoun; else N |
| 7 | NESTED_2 | Y if $m_k$ is a nested NP; else N |
| 8 | SEMCLASS_2 | the semantic class of $m_k$; can be one of PERSON, LOCATION, ORGANIZATION, DATE, TIME, MONEY, PERCENT, OBJECT, OTHERS, determined using WordNet (Fellbaum, 1998) and the Stanford NE recognizer (Finkel, Grenager, & Manning, 2005) |
| 9 | ANIMACY_2 | Y if $m_k$ is determined as HUMAN or ANIMAL by WordNet and an NE recognizer; else N |
| 10 | PRO_TYPE_2 | the nominative case of $m_k$ if it is a pronoun; else NA. E.g., the feature value for *him* is HE |

| **Features describing the relationship between $m_j$, a candidate antecedent and $m_k$, the mention to be resolved** | |
|---|---|
| 11 | HEAD_MATCH | C if the mentions have the same head noun; else I |
| 12 | STR_MATCH | C if the mentions are the same string; else I |
| 13 | SUBSTR_MATCH | C if one mention is a substring of the other; else I |
| 14 | PRO_STR_MATCH | C if both mentions are pronominal and are the same string; else I |
| 15 | PN_STR_MATCH | C if both mentions are proper names and are the same string; else I |
| 16 | NONPRO_STR_MATCH | C if the two mentions are both non-pronominal and are the same string; else I |
| 17 | MODIFIER_MATCH | C if the mentions have the same modifiers; NA if one of both of them don't have a modifier; else I |
| 18 | PRO_TYPE_MATCH | C if both mentions are pronominal and are either the same pronoun or different only with respect to case; NA if at least one of them is not pronominal; else I |
| 19 | NUMBER | C if the mentions agree in number; I if they disagree; NA if the number for one or both mentions cannot be determined |
| 20 | GENDER | C if the mentions agree in gender; I if they disagree; NA if the gender for one or both mentions cannot be determined |
| 21 | AGREEMENT | C if the mentions agree in both gender and number; I if they disagree in both number and gender; else NA |
| 22 | ANIMACY | C if the mentions match in animacy; I if they don't; NA if the animacy for one or both mentions cannot be determined |
| 23 | BOTH_PRONOUNS | C if both mentions are pronouns; I if neither are pronouns; else NA |
| 24 | BOTH_PROPER_NOUNS | C if both mentions are proper nouns; I if neither are proper nouns; else NA |
| 25 | MAXIMALNP | C if the two mentions does not have the same maximial NP projection; else I |
| 26 | SPAN | C if neither mention spans the other; else I |
| 27 | INDEFINITE | C if $m_k$ is an indefinite NP and is not in an appositive relationship; else I |
| 28 | APPOSITIVE | C if the mentions are in an appositive relationship; else I |
| 29 | COPULAR | C if the mentions are in a copular construction; else I |
| **Features describing the relationship between $m_j$, a candidate antecedent and $m_k$, the mention to be resolved (continued from the previous page)** | | |
| 30 | SEMCLASS | C if the mentions have the same semantic class (where the set of semantic classes considered here is enumerated in the description of the SEMCLASS_2 feature); I if they don't; NA if the semantic class information for one or both mentions cannot be determined |
| 31 | ALIAS | C if one mention is an abbreviation or an acronym of the other; else I |
| 32 | DISTANCE | binned values for sentence distance between the mentions |
| **Additional features describing the relationship between $m_j$, a candidate antecedent and $m_k$, the mention to be resolved** | | |
| 33 | NUMBER' | the concatenation of the NUMBER_2 feature values of $m_j$ and $m_k$. E.g., if $m_j$ is *Clinton* and $m_k$ is *they*, the feature value is SINGULAR-PLURAL, since $m_j$ is singular and $m_k$ is plural |
| 34 | GENDER' | the concatenation of the GENDER_2 feature values of $m_j$ and $m_k$ |
| 35 | PRONOUN' | the concatenation of the PRONOUN_2 feature values of $m_j$ and $m_k$ |
| 36 | NESTED' | the concatenation of the NESTED_2 feature values of $m_j$ and $m_k$ |
| 37 | SEMCLASS' | the concatenation of the SEMCLASS_2 feature values of $m_j$ and $m_k$ |
| 38 | ANIMACY' | the concatenation of the ANIMACY_2 feature values of $m_j$ and $m_k$ |
| 39 | PRO_TYPE' | the concatenation of the PRO_TYPE_2 feature values of $m_j$ and $m_k$ |

Table 1: Feature set for coreference resolution. Non-relational features describe a mention and in most cases take on a value of **Yes** or **No**. Relational features describe the relationship between the two mentions and indicate whether they are **Compatible**, **Incompatible** or **Not Applicable**.

value of $-1$. When used in *classification* mode, an SVM learner aims to learn a hyperplane (i.e., a linear classifier) that separates the positive points from the negative points. If there is more than one hyperplane that achieves zero training error, the learner will choose the hyperplane that maximizes the *margin* of separation (i.e., the distance between the hyperplane and the training example closest to it), as a larger margin can be proven to provide better generalization on unseen data (Vapnik, 1995). More formally, a maximum margin hyperplane is defined by $\mathbf{w} \cdot \mathbf{x} - b = 0$, where $\mathbf{x}$ is a feature vector representing an arbitrary data point, and $\mathbf{w}$ (a weight vector) and $b$ (a scalar) are parameters that are learned by solving the following constrained optimization problem:

OPTIMIZATION PROBLEM 1: HARD-MARGIN SVM FOR CLASSIFICATION

$$\arg\min \quad \frac{1}{2}\|\mathbf{w}\|^2$$

subject to

$$y_i(\mathbf{w} \cdot \mathbf{x}_i - b) \geq 1, \quad 1 \leq i \leq n,$$

where $y_i \in \{+1, -1\}$ is the class of the $i$-th training point $\mathbf{x}_i$. Note that for each data point $\mathbf{x}_i$, there is exactly one linear constraint in this optimization problem that ensures $\mathbf{x}_i$ is correctly classified. In particular, using a value of 1 on the right side of each inequality constraint ensures a certain distance (i.e., margin) between each $\mathbf{x}_i$ and the hyperplane. It can be shown that the margin is inversely proportional to the length of the weight vector. Hence, minimizing the length of the weight vector is equivalent to maximizing the margin. The resulting SVM classifier is known as a *hard-margin* SVM: the margin is "hard" because each data point has to be on the correct side of the hyperplane.

However, in cases where the data set is not linearly separable, there is no hyperplane that can perfectly separate the positives from the negatives, and as a result, the above constrained optimization problem does not have a solution. Instead of asking the SVM learner to give up and return no solution, we solve a relaxed version of the problem where we also consider hyperplanes that produce non-zero training errors as potential solutions. In other words, we have to modify the linear constraints associated with each data point so that training errors are allowed. However, if we only modify the linear constraints but leave the objective function as it is, then the learner will only search for a maximum-margin hyperplane regardless of the training error it produces. Since training error correlates positively with generalization error, it is crucial for the objective function to also take into consideration the training error so that a hyperplane with a large margin and a low training error can be found. However, it is non-trivial to maximize the margin and minimize the training error simultaneously, since training error typically increases as we maximize the margin. As a result, we need to find a trade-off between these two criteria, resulting in an objective function that is a linear combination of margin size and training error. More formally, we find the optimal hyperplane by solving the following constrained optimization problem:

Optimization Problem 2: Soft-Margin SVM for Classification

$$\arg\min \quad \frac{1}{2}\|\mathbf{w}\|^2 + C\sum_i \xi_i$$

subject to

$$y_i(\mathbf{w} \cdot \mathbf{x}_i - b) \geq 1 - \xi_i, \quad 1 \leq i \leq n.$$

As before, $y_i \in \{+1, -1\}$ is the class of the $i$-th training point $\mathbf{x}_i$. $C$ is a regularization parameter that balances training error and margin size. Finally, $\xi_i$ is a non-negative *slack variable* that represents the degree of misclassification of $\mathbf{x}_i$; in particular, if $\xi_i > 1$, then data point $i$ is on the wrong side of the hyperplane. Because this SVM allows data points to appear on the wrong side of the hyperplane, it is also known as a *soft-margin* SVM. Given this optimization problem, we rely on the training algorithm employed by SVM$^{light}$ for finding the optimal hyperplane.

After training, the resulting SVM classifier is used by a clustering algorithm to identify an antecedent for a mention in a test text. Specifically, each active mention is compared in turn to each preceding mention. For each pair, a test instance is created as during training and presented to the SVM classifier, which returns a value that indicates the likelihood that the two mentions are coreferent. Mention pairs with class values above 0 are considered coreferent; otherwise the pair is considered not coreferent. Following Soon et al. (2001), we apply a *closest-first* linking regime for antecedent selection: given an active mention $m_k$, we select as its antecedent the closest preceding mention that is classified as coreferent with $m_k$. If $m_k$ is not classified as coreferent with any preceding mention, it will be considered non-anaphoric (i.e., no antecedent will be selected for $m_k$).

## 3.2 Entity-Mention Model

Unlike the mention-pair model, the entity-mention model is a classifier that decides whether or not an active mention $m_k$ belongs to a *partial coreference cluster* $c_j$ that precedes $m_k$. Each training instance, $\boldsymbol{i}(c_j, m_k)$, represents $c_j$ and $m_k$. The features for an instance can be divided into two types: (1) features that describe $m_k$ (i.e, those shown in the second block of Table 1), and (2) cluster-level features, which describe the relationship between $c_j$ and $m_k$. A cluster-level feature can be created from a feature employed by the mention-pair model by applying a logical predicate. For example, given the NUMBER feature (i.e., feature #19 in Table 1), which determines whether two mentions agree in number, we can apply the ALL predicate to create a cluster-level feature that has the value YES if $m_k$ agrees in number with *all* of the mentions in $c_j$ and NO otherwise. Motivated by previous work (Luo et al., 2004; Culotta, Wick, & McCallum, 2007; Yang et al., 2008), we create cluster-level features from mention-pair features using four commonly-used logical predicates: NONE, MOST-FALSE, MOST-TRUE, and ALL. Specifically, for each feature X shown in the last two blocks in Table 1, we first convert X into an equivalent set of binary-valued features if it is multi-valued. Then, for each resulting binary-valued feature $X_b$, we create four binary-valued cluster-level features: (1) NONE-$X_b$ is true when $X_b$ is false between $m_k$ and each mention in $c_j$; (2) MOST-FALSE-$X_b$ is true when $X_b$ is true between $m_k$ and less than half (but at least one) of the mentions in $c_j$; (3) MOST-TRUE-$X_b$ is true when $X_b$ is true between $m_k$ and at least half (but not all) of the mentions in $c_j$; and (4) ALL-$X_b$ is true when $X_b$ is true between $m_k$ and each mention in $c_j$. Hence, for each $X_b$, exactly one of these four cluster-level features evaluates to true.[7]

Following Yang et al. (2008), we create (1) a positive instance for each anaphoric mention $m_k$ and the preceding cluster $c_j$ to which it belongs; and (2) a negative instance for $m_k$ paired with each preceding cluster whose last mention appears between $m_k$ and its closest antecedent (i.e., the last mention of $c_j$). Consider again our running example. Three training instances will be generated for *He*: $\boldsymbol{i}(\{Monday\}, He)$, $\boldsymbol{i}(\{secretary\ of\ state\}, He)$, and $\boldsymbol{i}(\{Barack\ Obama,\ his\}, He)$. The first two of these instances will be labeled as negative, and the last one will be labeled as positive. As in the mention-pair model, we train the entity-mention model using the SVM learner.

Since the entity-mention model is a classifier, we will again use SVM$^{light}$ in classification mode, resulting in a constrained optimization problem that is essentially the same as OPTIMIZATION PROBLEM 2, except that each training example $\mathbf{x}_i$ represents an active mention and one of its preceding clusters rather than two mentions.

---

7. Note that a cluster-level feature can also be represented as a *probabilistic* feature. Specifically, recall that the four logical predicates partitions the [0,1] interval. Which predicate evaluates to true for a given cluster-level feature depends on the probability obtained during the computation of the feature. Instead of applying the logical predicates to convert the probability into one of the four discrete values, we can simply use the probability as the value of the cluster-level feature. However, we choose not to employ this probabilistic representation, as preliminary experiments indicated that using probabilistic features yielded slightly worse results than using logical features.

After training, the resulting classifier is used to identify a preceding cluster for a mention in a test text. Specifically, the mentions are processed in a left-to-right manner. For each active mention $m_k$, a test instance is created between $m_k$ and each of the preceding clusters formed so far. All the test instances are then presented to the classifier. Finally, we adopt a closest-first clustering regime, linking $m_k$ to the closest preceding cluster that is classified as coreferent with $m_k$. If $m_k$ is not classified as coreferent with any preceding cluster, it will be considered non-anaphoric. Note that all partial clusters preceding $m_k$ are formed incrementally based on the predictions of the classifier for the first $k - 1$ mentions; no gold-standard coreference information is used in their formation.

### 3.3 Mention-Ranking Model

As noted before, a ranking model imposes a ranking on all the candidate antecedents of an active mention $m_k$. To train the ranking-model, we use the SVM ranker-learning algorithm from Joachims's (2002) SVM$^{light}$ package.

Like the mention-pair model, each training instance $\boldsymbol{i}(m_j, m_k)$ represents $m_k$ and a preceding mention $m_j$. In fact, the features that represent an instance and the method for creating training instances are identical to those employed by the mention-pair model. The only difference lies in labeling the training instances. Assuming that $S_k$ is the set of training instances created for anaphoric mention $m_k$, the rank value for $\boldsymbol{i}(m_j, m_k)$ in $S_k$ is the rank of $m_j$ among competing candidate antecedents, which is 2 if $m_j$ is the closest antecedent of $m_k$, and 1 otherwise.[8] To exemplify, consider again our running example. As in the mention-pair model, three training instances will be generated for *He*: $\boldsymbol{i}(Monday, He)$, $\boldsymbol{i}(secretary\ of\ state, He)$, $\boldsymbol{i}(his, He)$. The third instance will have a rank value of 2, and the remaining two will have a rank value of 1.

At first glance, it seems that the training set that is generated for learning the mention-ranking model, is identical to the one for learning the mention-pair model, as each instance represents two mentions and is labeled with one of two possible values. Since previous work on ranking-based coreference resolution does not attempt to clarify the difference between the two, we believe that it could be difficult for the reader to appreciate the idea of using ranking for coreference resolution.

Let us first describe the difference between classification and ranking at a high level, beginning with the training sets employed by the mention-ranking model and the mention-pair model. The difference is that the label associated with each instance for training the mention-ranking model is a *rank* value, whereas the label associated with each instance for training the mention-pair model is a *class* value. More specifically, since a *ranking* SVM learns to rank a set of candidate antecedents, it is the *relative* ranks between two candidates, rather than the absolute rank of a candidate, that matter in the training process. In other words, from the point of view of the ranking SVM, a training set where instance #1 has a rank value of 2 and instance #2 has a rank value of 1 is functionally equivalent to one where #1 has a rank value of 10 and #2 has a rank value of 5, assuming that the remaining instances generated for the same anaphor in the two training sets are identical to each other and do not have a rank value between 1 and 10.

---

8. A larger rank value implies a better rank in SVM$^{light}$.

Next, we take a closer look at the ranker-training process. We denote the training set that is created as described above by $T$. In addition, we assume that an instance in $T$ is denoted by $(x_{jk}, y_{jk})$, where $x_{jk}$ is the feature vector created from anaphoric mention $m_k$ and candidate antecedent $m_j$, and $y_{jk}$ is its rank value. Before training a ranker, the SVM ranker-learning algorithm derives a training set $T'$ from the original training set $T$ as follows. Specifically, for every pair of training instances $(x_{ik}, y_{ik})$ and $(x_{jk}, y_{jk})$ in $T$ where $y_{ik} \neq y_{jk}$, we create a new training instance $(x_{ijk}, y_{ijk})$ for $T'$, where $x_{ijk} = x_{ik} - x_{jk}$, and $y_{ijk} \in \{+1, -1\}$ is 1 if $x_{ik}$ has a larger rank value than $x_{jk}$ (and $-1$ otherwise). In a way, the creation of $T'$ resembles Connolly et al.'s (1994) pairwise ranking approach that we saw in Section 2, where we convert a ranking problem into a pairwise classification problem.[9] The goal of the ranker-learning algorithm, then, is to find a hyperplane that minimizes the number of misclassifications in $T'$. Note that since $y_{ijk} \in \{+1, -1\}$, the class value of an instance in $T'$ depends only on the relative ranks of two candidate antecedents, not their absolute rank values.

Given the conversion from a ranking problem to a pairwise classification problem, the constrained optimization problem that the SVM ranker-learning algorithm attempts to solve, as described below, is similar to Optimization Problem 2:

Optimization Problem 3: Soft-Margin SVM for Ranking

$$\arg\min \quad \frac{1}{2}\|\mathbf{w}\|^2 + C\sum \xi_{ijk}$$

subject to

$$y_{ijk}(\mathbf{w} \cdot (\mathbf{x}_{ik} - \mathbf{x}_{jk}) - b) \geq 1 - \xi_{ijk},$$

where $\xi_{ijk}$ is a non-negative slack variable that represents the degree of misclassification of $\mathbf{x}_{ijk}$, and $C$ is a regularization parameter that balances training error and margin size.

Two points deserve mention. First, this optimization problem is equivalent to the one for a classification SVM on pairwise difference feature vectors $\mathbf{x}_{ik} - \mathbf{x}_{jk}$. As a result, the training algorithm that was used to solve Optimization Problem 2 is also applicable to this optimization problem. Second, while the number of linear inequality constraints generated from document $d$ in the optimization problems for training the mention-pair model and the entity-mention model is quadratic in the number of mentions in $d$, the number of constraints generated for a ranking SVM is cubic in the number of mentions, since each instance now represents three (rather than two) mentions.

After training, the mention-ranking model is applied to rank the candidate antecedents for an active mention in a test text as follows. Given an active mention $m_k$, we follow Denis and Baldridge (2008) and use an independently-trained classifier to determine whether $m_k$ is non-anaphoric. If so, $m_k$ will not be resolved. Otherwise, we create test instances for $m_k$ by pairing it with each of its preceding mentions. The test instances are then presented to the ranker, which computes a rank value for each instance by taking the dot product of the instance vector and the weight vector. The preceding mention that is assigned the largest value by the ranker is selected as the antecedent of $m_k$. Ties are broken by preferring the antecedent that is closest in distance to $m_k$.

---

9. The main difference between $T'$ and the training set employed by Connolly et al.'s approach is that in $T'$, each instance is formed by taking the difference of the feature vectors of two instances in $T$, whereas in Connolly et al.'s training set, each instance is formed by concatenating the feature vectors of two instances in $T$.

The anaphoricity classifier used in the resolution step is trained using a publicly-available implementation[10] of maximum entropy (MaxEnt) modeling. Each instance corresponds to a mention and is represented by 26 features that are deemed useful for distinguishing between anaphoric and non-anaphoric mentions (see Table 2 for details). Linguistically, these features can be broadly divided into three types: string-matching, grammatical, and semantic. Each of them is either a *relational* feature, which compares a mention to one of its preceding mentions, or a *non-relational* feature, which encodes certain linguistic property of the mention whose anaphoricity is to be determined (e.g., NP type, number, definiteness).

## 4. Coreference as Cluster Ranking

In this section, we describe our cluster-ranking approach to NP coreference. As noted before, our approach aims to combine the strengths of the entity-mention model and the mention-ranking model.

### 4.1 Training and Applying a Cluster Ranker

For ease of exposition, we will describe in this subsection how to train and apply the cluster-ranking model when it is used in a pipeline architecture, where anaphoricity determination is performed prior to coreference resolution. In the next subsection, we will show how the two tasks can be learned jointly.

Recall that the cluster-ranking model ranks a set of preceding clusters for an active mention $m_k$. Since the cluster-ranking model is a hybrid of the mention-ranking model and the entity-mention model, the way it is trained and applied is also a hybrid of the two. In particular, the instance representation employed by the cluster-ranking model is identical to that used by the entity-mention model, where each training instance $\boldsymbol{i}(c_j, m_k)$ represents a preceding cluster $c_j$ and an anaphoric mention $m_k$ and consists of cluster-level features formed from predicates. Unlike in the entity-mention model, however, in the cluster-ranking model, (1) a training instance is created between each anaphoric mention $m_k$ and *each* of its preceding clusters; and (2) since we are training a model for ranking clusters, the assignment of rank values to training instances is similar to that of the mention-ranking model. Specifically, the rank value of a training instance $\boldsymbol{i}(c_j, m_k)$ created for $m_k$ is the rank of $c_j$ among the competing clusters, which is 2 if $m_k$ belongs to $c_j$, and 1 otherwise.

To train the cluster-ranking model, we use the SVM learner in ranking mode, resulting in a constrained optimization problem that is essentially the same as OPTIMIZATION PROBLEM 3, except that each training example $\mathbf{x}_{ijk}$ represents an active mention $m_k$ and two of its preceding clusters, $c_i$ and $c_j$, rather than two of its preceding mentions.

Applying the learned cluster ranker to a test text is similar to applying the mention-ranking model. Specifically, the mentions are processed in a left-to-right manner. For each active mention $m_k$, we first apply an independently-trained classifier to determine if $m_k$ is non-anaphoric. If so, $m_k$ will not be resolved. Otherwise, we create test instances for $m_k$ by pairing it with each of its preceding clusters. The test instances are then presented to the ranker, and $m_k$ is linked to the cluster that is assigned the highest value by the ranker. Ties are broken by preferring the cluster whose last mention is closest in distance to $m_k$. Note that these partial clusters preceding $m_k$ are formed incrementally based on the predictions of the ranker for the first $k - 1$ mentions.

---

10. See `http://homepages.inf.ed.ac.uk/s0450736/maxent_toolkit.html`.

| Feature Type | Feature | Description |
|---|---|---|
| Lexical | STR_MATCH | Y if there exists a mention $m_j$ preceding $m_k$ such that, after discarding determiners, $m_j$ and $m_k$ are the same string; else N. |
| | HEAD_MATCH | Y if there exists a mention $m_j$ preceding $m_k$ such that $m_j$ and $m_k$ have the same head; else N. |
| | UPPERCASE | Y if $m_k$ is entirely in uppercase; else N. |
| Grammatical (NP type) | DEFINITE | Y if $m_k$ starts with "the"; else N. |
| | DEMONSTRATIVE | Y if $m_k$ starts with a demonstrative such as "this", "that", "these", or "those"; else N. |
| | INDEFINITE | Y if $m_k$ starts with "a" or "an"; else N. |
| | QUANTIFIED | Y if $m_k$ starts with quantifiers such as "every", "some", "all", "most", "many", "much", "few", or "none"; else N. |
| | ARTICLE | DEFINITE if $m_k$ is a definite NP; QUANTIFIED if $m_k$ is a quantified NP; else INDEFINITE. |
| | PRONOUN | Y if $m_k$ is a pronoun; else N. |
| | PROPER_NOUN | Y if $m_k$ is a proper noun; else N. |
| | BARE_SINGULAR | Y if $m_k$ is singular and does not start with an article; else N. |
| | BARE_PLURAL | Y if $m_k$ is plural and does not start with an article; else N. |
| | EMBEDDED | Y if $m_k$ is a prenominal modifier; else N. |
| Grammatical (NP property/ relationship) | APPOSITIVE | Y if $m_k$ is the first of the two mentions in an appositive construction; else N. |
| | PREDNOM | Y if $m_k$ is the first of the two mentions in a predicate nominal construction; else N. |
| | NUMBER | SINGULAR if $m_k$ is singular in number; PLURAL if $m_k$ is plural in number; UNKNOWN if the number information cannot be determined. |
| | CONTAINS_PN | Y if $m_k$ is not a proper noun but contains a proper noun; else N. |
| Grammatical (Syntactic Pattern) | THE_N | Y if $m_k$ starts with "the" followed exactly by one common noun; else N. |
| | THE_2N | Y if $m_k$ starts with "the" followed exactly by two common nouns; else N. |
| | THE_PN | Y if $m_k$ starts with "the" followed exactly by a proper noun; else N. |
| | THE_PN_N | Y if $m_k$ starts with "the" followed exactly by a proper noun and a common noun; else N. |
| | THE_ADJ_N | Y if $m_k$ starts with "the" followed exactly by an adjective and a common noun; else N. |
| | THE_NUM_N | Y if $m_k$ starts with "the" followed exactly by a cardinal and a common noun; else N. |
| | THE_NE | Y if $m_k$ starts with "the" followed exactly by a named entity; else N. |
| | THE_SING_N | Y if $m_k$ starts with "the" followed by a singular NP not containing any proper noun; else N. |
| Semantic | ALIAS | Y if there exists a mention $m_j$ preceding $m_k$ such that $m_j$ and $m_k$ are aliases; else N. |

Table 2: Feature set for anaphoricity determination. Each instance represents a single mention, $m_k$, characterized by 26 features.

## 4.2 Joint Anaphoricity Determination and Coreference Resolution

The cluster ranker described above can be used to determine which preceding cluster an anaphoric mention should be linked to, but it cannot be used to determine whether a mention is anaphoric or not. The reason is simple: all the training instances are generated from anaphoric mentions. Hence, to jointly learn anaphoricity determination and coreference resolution, we must train the ranker using instances generated from *both* anaphoric and non-anaphoric mentions.

Specifically, when training the ranker, we provide each active mention with the option to start a new cluster by creating an additional instance that (1) contains features that solely describe the active mention (i.e., the features shown in the second block of Table 1), and (2) has the highest rank value among competing clusters (i.e., 2) if it is non-anaphoric and the lowest rank value (i.e., 1) otherwise. The main advantage of jointly learning the two tasks is that it allows the ranking model to evaluate *all* possible options for an active mention (i.e., whether to resolve it, and if so, which preceding cluster is the best) *simultaneously*. Essentially the same method can be applied to jointly learn the two tasks for the mention-ranking model.

After training, the resulting cluster ranker processes the mentions in a test text in a left-to-right manner. For each active mention $m_k$, we create test instances for it by pairing it with each of its preceding clusters. To allow for the possibility that $m_k$ is non-anaphoric, we create an additional test instance that contains features that solely describe the active mention (similar to what we did in the training step above). All these test instances are then presented to the ranker. If the additional test instance is assigned the highest rank value by the ranker, then $m_k$ is classified as non-anaphoric and will not be resolved. Otherwise, $m_k$ is linked to the cluster that has the highest rank, with ties broken by preferring the antecedent that is closest to $m_k$. As before, all partial clusters preceding $m_k$ are formed incrementally based on the predictions of the ranker for the first $k - 1$ mentions.

Finally, we note that our model for jointly *learning* anaphoricity determination and coreference resolution is different from recent attempts to perform joint *inference* for anaphoricity determination and coreference resolution using integer linear programming (ILP), where an anaphoricity classifier and a coreference classifier are trained *independently* of each other, and then ILP is applied as a postprocessing step to jointly infer anaphoricity and coreference decisions so that they are consistent with each other (e.g., Denis & Baldridge, 2007a). Joint inference is different from our joint-learning approach, which allows the two tasks to be learned jointly and not independently.

## 5. Lexicalization for Coreference Resolution

Next, we investigate the role of lexicalization (i.e., the use of *word pairs* as linguistic features) in learning-based coreference resolution. The motivation behind our investigation is two-fold. First, lexical features are very easy to compute and yet they are under-investigated in coreference resolution. In particular, only a few attempts have been made to employ them to train the mention-pair model (e.g., Luo et al., 2004; Daumé III & Marcu, 2005; Bengtson & Roth, 2008). In contrast, we want to determine whether they can improve the performance of our cluster-ranking model. Second, the mention-pair model and the mention-ranking model have only been compared with respect to a non-lexical feature set (Denis & Baldridge, 2007b, 2008), so it is not clear how they will perform relative to each other when they are trained on lexical features. We desire an answer to this question, as it will allow us to gain additional insights into the strengths and weaknesses of these learning-based coreference models.

Recall from the introduction that previous attempts on lexicalizing the mention-pair model show that lexical features are at best marginally useful. Hence, one of our goals here is to determine whether we can make better use of lexical features for a learning-based coreference resolver. In particular, unlike the aforementioned attempts on lexicalization, which simply append all word pairs to a "conventional" coreference feature set consisting of string-matching, grammatical, semantic, and distance (i.e., proximity-based) features (e.g., the feature set shown in Table 1), we investigate a model that exploits lexical features in combination with only a small subset of these conventional coreference features. This would allow us to have a better understanding of the significance of these conventional features. For example, features that encode agreement on gender, number, and semantic class between two mentions are employed by virtually all learning-based coreference resolver, but we never question whether there are better alternatives to these features. If we could build a lexicalized coreference model without these commonly-used features and did not observe any performance deterioration, it would imply that these conventional features were replaceable, and that there was no prototypical way of building a learning-based coreference system.

The question is: what is the small subset of conventional features that we should use in combination with the lexical features? As mentioned above, since one of the advantages of lexical features is that they are extremely easy to compute, we desire only those conventional features that are also easy to compute, especially those that do *not* require a dictionary to compute. As we will see, we choose to use only two features, the ALIAS feature and the DISTANCE feature (see features 31 and 32 in Table 1), and rely on an off-the-shelf named entity (NE) recognizer to compute NE types.

Note, however, that the usefulness of lexical features could be limited in part by data sparseness: many word pairs that appear in the training data may not appear in the test data. While employing some of the conventional features described above (e.g., DISTANCE) will help alleviate this problem, we seek to further improve generalizability by introducing two types of features: *semi-lexical* and *unseen* features. We will henceforth refer to the feature set that comprises these two types of features, the lexical features, the ALIAS feature, and the DISTANCE feature as the *Lexical* feature set. In addition, we will refer to the feature set shown in Table 1 as the *Conventional* feature set.

Below we first describe the *Lexical* feature set for training the mention-pair model and the mention-ranking model (Section 5.1). After that, we show how to create cluster-level features from this feature set for training the entity-mention model and the cluster-ranking model, as well as issues in training a joint model for anaphoricity determination and coreference resolution (Section 5.2).

## 5.1 Lexical Feature Set

Unlike previous work on lexicalizing learning-based coreference models, our Lexical feature set consists of four types of features: lexical features, semi-lexical features, unseen features, as well as two "conventional" features (namely, ALIAS and DISTANCE).

To compute these features, we preprocess a *training* text by randomly replacing 10% of its nominal mentions (i.e., common nouns) with the label UNSEEN. If a mention $m_k$ is replaced with UNSEEN, all mentions that have the same string as $m_k$ will also be replaced with UNSEEN. A *test* text is preprocessed differently: we simply replace all mentions whose strings are not seen in the training data with UNSEEN. Hence, artificially creating UNSEEN labels from a training text will allow a learner to learn how to handle unseen words in a test text, potentially improving generalizability.

After preprocessing, we can compute the features for an instance. Assuming that we are training the mention-pair model or the mention-ranking model, each instance corresponds to two mentions, $m_j$ and $m_k$, where $m_j$ precedes $m_k$ in the text. The features can be divided into four groups: unseen, lexical, semi-lexical, and conventional. Before describing these features, two points deserve mention. First, if at least one of $m_j$ and $m_k$ is UNSEEN, no lexical, semi-lexical, or conventional features will be created for them, since features involving an UNSEEN mention are likely to be misleading for a learner in the sense that they may yield incorrect generalizations from the training set. Second, since we use an SVM for training and testing, each instance can contain any number of features, and unless otherwise stated, a feature has the value 1.

**Unseen feature.** If both $m_j$ and $m_k$ are UNSEEN, we determine whether they are the same string. If so, we create an UNSEEN-SAME feature; otherwise, we create an UNSEEN-DIFF feature. If only one of them is UNSEEN, no feature will be created.

**Lexical feature.** We create a lexical feature between $m_j$ and $m_k$, which is an ordered pair consisting of the heads of the mentions. For a pronoun or a common noun, the head is assumed to be the last word of the mention[11]; for a proper noun, the head is taken to be the entire noun phrase.

**Semi-lexical features.** These features aim to improve generalizability. Specifically, if exactly one of $m_j$ and $m_k$ is tagged as an NE by the Stanford NE recognizer (Finkel et al., 2005), we create a semi-lexical feature that is identical to the lexical feature described above, except that the NE is replaced with its NE label (i.e., PERSON, LOCATION, ORGANIZATION). If both mentions are NEs, we check whether they are the same string. If so, we create the feature *NE*-SAME, where *NE* is replaced with the corresponding NE label. Otherwise, we check whether they have the same NE tag *and* a word-subset match (i.e., whether all word tokens in one mention appear in the other's list of tokens). If so, we create the feature *NE*-SUBSAME, where *NE* is replaced with their NE label. Otherwise, we create a feature that is the concatenation of the NE labels of the two mentions.

---

11. As we will see in the evaluation section, our mention extractor is trained to extract base NPs. Hence, while our heuristic for extracting head nouns is arguably overly simplistic, it will not be applied to recursive NPs (e.g., NPs that contain prepositional phrases), which are phrases on which it is likely to make mistakes. However, if we desire a better extraction accuracy, we can extract the head nouns from syntactic parsers that provide head information, such as Collins's (1999) parser.

**Conventional features.** To further improve generalizability, we incorporate two easy-to-compute features from the *Conventional* feature set: ALIAS and DISTANCE.

## 5.2 Feature Generation

Now that we have a *Lexical* feature set for training the mention-pair model and the mention-ranking model, we can describe the two extensions to this feature set that are needed to (1) train the entity-mention model and the cluster-ranking model, and (2) perform joint learning for anaphoricity determination and coreference resolution.

The first extension concerns the generation of cluster-level features for the entity-mention model and the cluster-level model. Recall from Section 3.2 that to create cluster-level features given the *Conventional* feature set, we first convert each feature employed by the mention-pair model into an equivalent set of binary-valued features, and then create a cluster-level feature from each of the resulting binary-valued features. On the other hand, given the *Lexical* feature set, this method of producing cluster-level features is only applicable to the two "conventional" features (i.e., ALIAS and DISTANCE), as they also appear in the *Conventional* feature set. For an unseen, lexical, or semi-lexical feature, we create a feature between the active mention and *each* mention in the preceding cluster, as described in Section 5.1[12], and the value of this feature is the number of times it appears in the instance. Encoding feature values as frequency rather than binary values allows us to capture cluster-level information in a shallow manner.

The second extension concerns the generation of features for representing the additional instance that is created when training the joint version of the mention-ranking model and the cluster-ranking model. Recall from Section 4.2 that when the *Conventional* feature set was used, we represented this additional instance using features that were computed solely from the active mention. On the other hand, given the *Lexical* feature set, we can no longer use the same method for representing this additional instance, as there is no feature in the *Lexical* feature set that is computed solely from the active mention. As a result, we represent this additional instance using just one feature, NULL-X, where X is the head of the active mention, to help the learner learn that X is likely to be non-anaphoric.

## 6. Evaluation

Our evaluation is driven by the following questions, focusing on (1) the comparison among different learning-based coreference models, and (2) the effect of lexicalization on these models. Specifically:

- How do the learning-based coreference models (namely, the mention-pair model, the entity-mention model, the mention-ranking model, and our cluster-ranking model) compare with each other?

---

12. Strictly speaking, the resulting feature is *not* a cluster-level feature, as it is computed between an active mention and only *one* of the mentions in the preceding cluster.

- Does joint modeling for anaphoricity determination and coreference resolution offer any benefits over the pipeline architecture, where anaphoricity is performed prior to coreference resolution?

- Do lexicalized coreference models perform better than their unlexicalized counterparts?

In the rest of this section, we will first describe the experimental setup (Section 6.1), and then show the performance of the four models, including the effect of lexicalization and joint modeling whenever applicable, on three different feature sets (Section 6.2).

## 6.1 Experimental Setup

We begin by providing the details on the data sets, our automatic mention extraction method, and the scoring programs.

### 6.1.1 Corpus

We use the ACE 2005 coreference corpus as released by the LDC, which consists of the 599 training documents used in the official ACE evaluation.[13] To ensure diversity, the corpus was created by selecting documents from six different sources: Broadcast News (BN), Broadcast Conversations (BC), Newswire (NW), Webblog (WB), Usenet (UN), and Conversational Telephone Speech (CTS). The number of documents belonging to each source is shown in Table 3.

| Data set | BN | BC | NW | WL | UN | CTS |
|---|---|---|---|---|---|---|
| # of documents | 226 | 60 | 106 | 119 | 49 | 39 |

Table 3: Statistics for the ACE 2005 corpus

### 6.1.2 Mention Extraction

We evaluate each coreference model using *system mentions*. To extract system mentions from a test text, we trained a mention extractor on the training texts. Following Florian et al. (2004), we recast mention extraction as a sequence labeling task, where we assign to each token in a test text a label that indicates whether it **b**egins a mention, is **i**nside a mention, or is **o**utside a mention. Hence, to learn the extractor, we create one training instance for each token in a training text and derive its class value (one of **b**, **i**, and **o**) from the annotated data. Each instance represents $w_i$, the token under consideration, and consists of 29 linguistic features, many of which are modeled after the systems of Bikel, Schwartz, and Weischedel (1999) and Florian et al. (2004), as described below.

**Lexical (7):** Tokens in a window of 7: $\{w_{i-3}, \dots, w_{i+3}\}$.

**Capitalization (4):** Determine whether $w_i$ `IsAllCap`, `IsInitCap`, `IsCapPeriod`, and `IsAllLower`.

---

13. Since we did not participate in ACE 2005, we do not have access to the official test set.

**Morphological (8):** $w_i$'s prefixes and suffixes of length one, two, three, and four.

**Grammatical (1):** The part-of-speech (POS) tag of $w_i$ obtained using the Stanford log-linear POS tagger (Toutanova, Klein, Manning, & Singer, 2003).

**Semantic (1):** The named entity (NE) tag of $w_i$ obtained using the Stanford CRF-based NE recognizer (Finkel et al., 2005).

**Dictionaries (8):** We employ eight dictionary-based features that indicate the presence or absence of $w_i$ in a particular dictionary. The eight dictionaries contain pronouns (77 entries), common words and words that are not names (399.6k), person names (83.6k), person titles and honorifics (761), vehicle words (226), location names (1.8k), company names (77.6k), and nouns extracted from WordNet that are hyponyms of PERSON (6.3k).

We employ CRF++[14], a C++ implementation of conditional random fields, for training the mention detector on the training set. Overall, the detector achieves an F-measure of 86.7 (86.1 recall, 87.2 precision) on the test set. These extracted mentions are to be used as system mentions in our coreference experiments.

### 6.1.3 Scoring Programs

To score the output of a coreference model, we employ two scoring programs, B$^3$ (Bagga & Baldwin, 1998) and $\phi_3$-CEAF[15] (Luo, 2005), which address the inherent weaknesses of the MUC scoring program (Vilain, Burger, Aberdeen, Connolly, & Hirschman, 1995).[16] Both B$^3$ and CEAF score a *response* (i.e., system-generated) partition, $R$, against a *key* (i.e., gold-standard) partition, $K$, and report coreference performance in terms of recall, precision, and F-measure. B$^3$ first computes recall and precision for each mention, $m_k$, as follows:

$$recall(m_k) = \frac{|R_{m_k} \cap K_{m_k}|}{|K_{m_k}|}, \ \ precision(m_k) = \frac{|R_{m_k} \cap K_{m_k}|}{|R_{m_k}|},$$

where $R_{m_k}$ is the coreference cluster containing $m_k$ in $R$, and $K_{m_k}$ is the coreference cluster containing $m_k$ in $K$. Then it computes overall recall (resp. precision) by averaging the per-mention recall (resp. precision) scores.

On the the hand, CEAF first constructs the optimal one-to-one mapping between the clusters in the key partition and those in the response partition. Specifically, assume that $K = \{K_1, K_2, \ldots, K_m\}$ is the set of clusters in the key partition, and $R = \{R_1, R_2, \ldots, R_n\}$ is the set of clusters in the response partition. To compute recall, CEAF first computes the score of each cluster, $K_i$, in $K$ as follows:

$$score(K_i) = |K_i \cap R_j|,$$

---

14. Available from `http://crfpp.sourceforge.net`
15. CEAF has two versions: $\phi_3$-CEAF and $\phi_4$-CEAF. The two versions differ in how the similarity of two aligned clusters is computed. We refer the reader to Luo's (2005) paper for details. $\phi_3$-CEAF is chosen here because it is the more commonly-used version of CEAF.
16. Briefly, the MUC scoring program suffers from two often-cited weaknesses. First, as a *link-based* measure, it does not reward successful identification of singleton clusters, since the mentions in these clusters are not linked to any other mentions. Second, it tends to under-penalize partitions with overly large clusters. See the work of Bagga and Baldwin (1998), Luo (2005), and Recasens and Hovy (2011) for details.

where $R_j$ is the cluster to which $K_i$ is mapped in the optimal one-to-one mapping, which can be constructed efficiently using the Kuhn-Munkres algorithm (Kuhn, 1955). Note that if $K_i$ is not mapped to any cluster in $R$, then $score(K_i) = 0$. CEAF then computes recall by summing the score of each cluster in $K$ and dividing the sum by the number of mentions in $K$. Precision can be computed in the same manner, except that we reverse the roles of $K$ and $R$.

A complication arises when $B^3$ is used to score a response partition containing system mentions. Recall that $B^3$ constructs a mapping between the mentions in the response and those in the key. Hence, if the response is generated using gold-standard mentions, then every mention in the response is mapped to some mention in the key and vice versa. In other words, there are no *twinless* (i.e., unmapped) mentions (Stoyanov et al., 2009). This is not the case when system mentions are used, but the original description of $B^3$ does not specify how twinless mentions should be scored (Bagga & Baldwin, 1998). To address this problem, we set the per-mention recall and precision of a twinless mention to zero, regardless of whether the mention appears in the key or the response. Note that CEAF can compare partitions with twinless mentions without any modification, since it operates by aligning clusters, not mentions.

Additionally, we apply a preprocessing step to a response partition before scoring it: we remove all and only those twinless system mentions that are singletons. The reason is simple: since the coreference resolver has successfully identified these mentions as singletons, it should not be penalized, and removing them allows us to avoid such penalty. Note that we only remove twinless (as opposed to all) system mentions that are singletons: this allows us to reward a resolver for successful identification of singleton mentions that have twins. On the other hand, we retain (1) twinless system mentions that are non-singletons (as the resolver should be penalized for identifying spurious coreference relations) and (2) twinless mentions in the key partition (as we want to ensure that the resolver makes the correct coreference or non-coreference decisions for them).[17]

## 6.2 Results

Before showing the results of the learning-based coreference models, let us consider the "head match" baseline, which is a commonly-used heuristic baseline for coreference resolution. It posits two mentions as coreferent if and only if their head nouns match. Head nouns are determined as described in Section 5.1: the head of a proper noun is the string of the entire mention, whereas the head of a pronoun or a common noun is the last word of the mention. Since one of our goals is to examine the effect of lexicalization on a coreference model, the head match baseline can provide information on how well we can do with one of the simplest kinds of string matching. Results of this baseline, shown in row 1 of Table 4, are expressed in terms of recall (R), precision (P), and F-measure (F) obtained via $B^3$ and CEAF. As we can see from Table 4, this baseline achieves F-measure scores of 54.9 and 49.6 according to $B^3$ and CEAF, respectively.

---

17. In addition to the method described here, a number of methods have been proposed to address the mapping problem. We refer the reader to the work of Enrique, Gonzalo, Artiles, and Verdejo (2009), Stoyanov et al. (2009), and Cai and Strube (2010) for details.

Next, we train and evaluate the learning-based coreference models using five-fold cross validation. For each data set $s_i$ shown in Table 3, we partition the documents in $s_i$ into five folds of approximately equal size, $s_{i1}, \ldots, s_{i5}$. We then train each coreference model on four folds and use it to generate coreference chains for the documents in the remaining fold, repeating this step five times so that each fold is used as the test fold exactly once. After that, we apply $B^3$ and CEAF to the entire set of automatically coreference-annotated documents to obtain the scores in Table 4. Below we discuss the results of the learning-based coreference models obtained when used in combination with three feature sets: the *Conventional* feature set (Section 6.2.1), the *Lexical* feature set (Section 6.2.2), and the *Combined* feature set, which is composed of all the features from *Conventional* and *Lexical* (Section 6.2.3).

### 6.2.1 Results Using the Conventional Features

To gauge the performance of our cluster-ranking model, we employ as baselines the mention-pair model, the entity-mention model, and the mention-ranking model.

**The mention-pair baseline.** We train our first learning-based baseline, the mention-pair model, using the SVM learning algorithm as implemented in the SVM$^{light}$ package.[18] As we can see from row 2 of Table 4, the mention-pair model achieves F-measure scores of 58.6 ($B^3$) and 54.4 (CEAF), which represent a statistically significant improvement of 3.7% and 4.8% in F-measure over the corresponding results for the head match baseline.[19]

**The entity-mention baseline.** Next, we train our second learning-based baseline, the entity-mention model, using the SVM learner. As we can see from row 3 of Table 4, this baseline achieves F-measure scores of 58.9 ($B^3$) and 54.8 (CEAF), which represent small but statistically significant improvements over the mention-pair model. The significant performance difference is perhaps not particularly surprising given the improved expressiveness of the entity-mention model over the mention-pair model.

**The mention-ranking baseline.** Our third baseline is the mention-ranking model, which is trained using the ranker-learning algorithm in SVM$^{light}$. To identify non-anaphoric mentions, we employ two methods. In the first method, we follow Denis and Baldridge (2008) and adopt a pipeline architecture, where we train a MaxEnt classifier for anaphoricity determination independently of the mention ranker on the training set using the 26 features described in Section 3.3. We then apply the resulting classifier to each test text to filter non-anaphoric mentions prior to coreference resolution. Results of this pipeline mention ranker are shown in row 4 of Table 4. As we can see, the ranker achieves F-measure scores of 57.7 ($B^3$) and 53.0 (CEAF), yielding a significant performance deterioration in comparison to the entity-mention baseline.

In the second method, we perform anaphoricity determination jointly with coreference resolution using the method described in Section 4.2. While we discussed this joint learning method in the context of cluster ranking, it should be easy to see that the method is equally applicable to the mention-ranking model. Results of the mention ranker using this

---

18. For this and subsequent uses of the SVM learner, we set all parameters to their default values. In particular, we employ a linear kernel to obtain all the results in this article.
19. All statistical significance results in this article are obtained using the paired *t*-test, with $p < 0.05$.

| | Coreference Model | $B^3$ | | | CEAF | | |
|---|---|---|---|---|---|---|---|
| | | R | P | F | R | P | F |
| 1 | Head match | 44.1 | 72.9 | 54.9 | 41.9 | 60.8 | 49.6 |

Using the *Conventional* feature set

| | | | | | | | |
|---|---|---|---|---|---|---|---|
| 2 | Mention-pair model | 49.7 | 71.4 | 58.6 | 49.5 | 60.5 | 54.4 |
| 3 | Entity-mention model | 49.9 | 71.7 | 58.9 | 51.0 | 59.2 | 54.8 |
| 4 | Mention-ranking model (Pipeline) | 48.1 | 72.1 | 57.7 | 51.7 | 54.4 | 53.0 |
| 5 | Mention-ranking model (Joint) | 49.1 | 76.1 | 59.7 | 52.9 | 59.2 | 55.9 |
| 6 | Cluster-ranking model (Pipeline) | 49.9 | 71.6 | 58.8 | 53.4 | 54.6 | 54.0 |
| 7 | Cluster-ranking model (Joint) | 51.1 | 73.3 | **60.2** | 54.1 | 60.2 | **57.0** |

Using the *Lexical* feature set

| | | | | | | | |
|---|---|---|---|---|---|---|---|
| 8 | Mention-pair model | 53.0 | 75.3 | 62.2 | 55.6 | 62.0 | 58.6 |
| 9 | Entity-mention model | 53.1 | 75.8 | 62.5 | 55.7 | 62.2 | 58.8 |
| 10 | Mention-ranking model (Pipeline) | 55.7 | 67.5 | 61.0 | 56.3 | 62.0 | 59.0 |
| 11 | Mention-ranking model (Joint) | 56.6 | 73.1 | **63.8** | 58.5 | 65.0 | **61.6** |
| 12 | Cluster-ranking model (Pipeline) | 51.0 | 67.1 | 58.0 | 53.1 | 55.3 | 54.1 |
| 13 | Cluster-ranking model (Joint) | 51.3 | 75.7 | 61.1 | 53.3 | 58.6 | 55.8 |

Using the *Combined* feature set

| | | | | | | | |
|---|---|---|---|---|---|---|---|
| 14 | Mention-pair model | 50.4 | 73.1 | 59.6 | 53.8 | 61.9 | 57.5 |
| 15 | Entity-mention model | 50.5 | 73.4 | 59.8 | 54.1 | 62.3 | 57.9 |
| 16 | Mention-ranking model (Pipeline) | 50.6 | 74.6 | 60.3 | 54.1 | 61.8 | 57.7 |
| 17 | Mention-ranking model (Joint) | 49.9 | 79.3 | 61.3 | 54.7 | 61.4 | 57.9 |
| 18 | Cluster-ranking model (Pipeline) | 52.9 | 70.9 | 60.6 | 57.5 | 62.1 | 59.7 |
| 19 | Cluster-ranking model (Joint) | 54.3 | 75.1 | **63.0** | 57.6 | 64.3 | **60.8** |

Table 4: Five-fold cross-validation coreference results obtained using $B^3$ and CEAF. The best F-measure achieved for each feature set/scoring program combination is boldfaced.

joint architecture are shown in row 5 of Table 4. As we can see, the ranker achieves F-measure scores of 59.7 ($B^3$) and 55.9 (CEAF), which represent significant improvements over the entity-mention model and its pipeline counterpart. Not only do these results demonstrate the superiority of the joint mention-ranking model to the entity-mention model, they substantiate the hypothesis that joint modeling offers benefits over pipeline modeling.

**Our cluster-ranking model.** Finally, we evaluate our cluster-ranking model. As in the mention-ranking baselines, we employ both the pipeline architecture and the joint architecture for anaphoricity determination. Results are shown in rows 6 and 7 of Table 4, respectively, for the two architectures. As we can see, the pipeline architecture yields F-measure scores of 58.8 ($B^3$) and 54.0 (CEAF), which represent a significant improvement over the mention ranker adopting the pipeline architecture. With the joint architecture, the cluster ranker achieves F-measure scores of 60.2 ($B^3$) and 57.0 (CEAF). This also rep-resents a significant improvement over the mention ranker adopting the joint architecture, the best of the baselines. Taken together, these results demonstrate the superiority of the cluster ranker to the mention ranker. Finally, the fact that the joint cluster ranker performs significantly better than its pipeline counterpart provides further empirical support for the benefits of joint modeling over pipeline modeling.

### 6.2.2 Results Using the Lexical Features

Next, we evaluate the learning-based coreference models using the *Lexical* features. Results are shown in rows 8–13 of Table 4. In comparison to the results obtained using the *Conventional* features, we see a different trend: the joint mention-ranking model replaces the cluster-ranking model as the best-performing model. Moreover, its improvement over the second best-performing model, which is the entity-mention model according to $B^3$ and the pipeline mention-ranking model according to CEAF, is statistically significant regardless of which scoring program is used. A closer examination of the results reveals that employing *Lexical* rather than *Conventional* features substantially improves the performance of the mention-ranking model: in comparison to the unlexicalized joint mention-ranking model (row 5), the F-measure scores of the lexicalized joint mention-ranking model (row 11) rise by 4.1% ($B^3$) and 5.7% (CEAF). This increase in F-measure can be attributed primarily to a substantial rise in recall, even though there is also a large increase in CEAF precision. Besides the joint mention-ranking model, the mention-pair model and the entity-mention model also benefit substantially when the *Conventional* features are replaced with the *Lexical* features: we see that the F-measure scores increase by 3.6% ($B^3$) and 4.2% (CEAF) for the mention-pair model, and by 3.6% ($B^3$) and 4.0% (CEAF) for the entity-mention model. The gains in F-measure for these two models can be attributed to large increases in both recall and precision. On the other hand, the joint cluster-ranking model does not always improve when we replace the *Conventional* features with the *Lexical* features. In fact, the performance difference between the cluster-ranking model and the entity-mention model is statistically indistinguishable. Finally, we see the benefits of jointly learning anaphoricity determination and coreference resolution again: when the joint version of the mention-ranking model is used rather than the pipeline version (compare rows 10 and 11), the F-measure scores rise significantly by 2.8% ($B^3$) and 2.6% (CEAF). Similarly for the cluster-ranking model: the joint version improves the pipeline version significantly by 3.1% ($B^3$) and 1.7% (CEAF) in F-measure.

Overall, these results are somewhat unexpected: recall that the *Lexical* features are very knowledge-lean, consisting of lexical, semi-lexical, and unseen features, as well as only two *Conventional* features. In particular, it does *not* employ any conventional coreference features that encode agreement on gender and number. This implies that many existing implementations of the mention-pair model, the entity-mention model, and the mention-ranking model, which are unlexicalized and rely heavily on the conventional features, are not making effective use of the labeled data. Perhaps more importantly, our results indicate that these coreference models can perform well (and in fact better) even without the conventional coreference features. Since all of the *Lexical* can be computed extremely easily, they can readily be applied to other languages, which is another advantage of this feature set. On the other hand, it is interesting to see that both versions of the cluster-ranking model exhibit less dramatic changes in performance as we replace the *Conventional* features with the *Lexical* features.

### 6.2.3 RESULTS USING THE COMBINED FEATURES

Since the *Conventional* features and the *Lexical* features represent two fairly different sources of knowledge, we examine whether we can improve the coreference models by combining these two feature sets. Results of the coreference models using the *Combined* features are shown in rows 14–19 of Table 4. These results exhibit essentially the same trend as those that we obtained with the *Conventional* features, with the joint cluster-ranking model performing the best and the mention-pair model performing the worst. In fact, the joint cluster-ranking model yields significantly better performance when used with the *Combined* features than with the *Conventional* features or the *Lexical* features alone. Similarly for the pipeline cluster-ranking model, which achieves significantly better performance with the *Combined* features than with the *Conventional* or *Lexical* features. These results seem to suggest that the cluster-ranking model is able to exploit the potentially different sources of information provided by the two feature sets to improve its performance. In addition, they demonstrate the benefits of joint modeling: for the mention-ranking model, the joint version improves the pipeline version significantly by 1.0% ($B^3$) and 0.2% (CEAF) in F-measure; and for the cluster-ranking model, the joint version improves its pipeline counterpart significantly by 2.4% ($B^3$) and 1.1% (CEAF) in F-measure.

The remaining coreference models all exhibit a drop in performance when the *Combined* features are used in lieu of the *Lexical* features. These results seem to suggest that the cluster-ranking model offers more robust performance in the face of changes in the underlying feature set than the other coreference models, and that feature selection, an issue that is under-explored in coreference resolution, may be crucial when we employ the other coreference models.[20] Perhaps more importantly, despite the fact that the *Conventional* features and the *Lexical* features represent two fairly different sources of information, all but the cluster-ranking model are unable to exploit the potentially richer amount of information contained in the *Combined* feature set. Hence, while virtually all the linguistic features that are recently developed for supervised coreference resolution have been evaluated using the mention-pair model (see, for example, the work of Strube, Rapp, & Müller, 2002; Ji, Westbrook, & Grishman, 2005; Ponzetto & Strube, 2006), the utility of these features may be better demonstrated using the cluster-ranking model.

A natural question is: how does our joint cluster-ranking model compare to the existing coreference systems? Since we did not participate in the ACE evaluations, we do not have access to the official test sets with which we can compare our model against the ACE participating coreference systems. The comparison is further complicated by the fact that existing coreference systems have been evaluated on different data sets, including the two MUC data sets (MUC-6, 1995; MUC-7, 1998) and the various ACE data sets (e.g., ACE-2, ACE 2003, ACE 2004, ACE 2005), as well as on different partitions of a given data set. To our knowledge, the only coreference model that has been evaluated on the same test data as ours is Haghighi and Klein's (2010) unsupervised coreference model. Their model has recently been shown to surpass the performance of Stoyanov et al.'s (2009) system, which is one of the best existing implementations of the mention-pair model. On our test data, Haghighi and Klein's model achieves a $B^3$ F-measure of 62.7, while ours achieves a $B^3$ F-measure of 62.8.[21] These results provide suggestive evidence that our cluster-ranking model achieves performance that is comparable with one of the best existing coreference models.

---

20. In fact, Ng and Cardie (2002b), Strube and Müller (2003), and Ponzetto and Strube (2006) show that the mention-pair model can be improved using feature selection.

Nevertheless, we caution that these results do not allow one to claim anything more than the fact that our model compares favorably to Haghighi and Klein's (2010) model. For instance, one cannot claim that their model is better because it achieves the same level of performance as ours without using any labeled data. The reasons are that (1) the mentions used by the two models in the coreference process are extracted differently and (2) the linguistic features employed by the two models and the way these features are computed are also different from each other. Since previous work has shown that these linguistic preprocessing steps can have a considerable impact on the performance of a resolver (Barbu & Mitkov, 2001; Stoyanov et al., 2009), it is possible that if one model employed the features or the mentions that the other model is currently using, then the results would be different. Hence, if one is to fairly compare two coreference models, they should be evaluated on the same set of mentions (rather than just the same set of documents) and are given access to the same set of knowledge sources, in essentially the same way as we compare the various learning-based coreference models in this article.

## 7. Experimental Analyses

In an attempt to gain insights into the different aspects of our coreference models, we conduct additional experiments and analyses. Rather than report five-fold cross-validation results, in this section we report results on one fold (i.e., the fold we designate as the test set) and use the remaining four folds solely for training.

### 7.1 Improving Classification-Based Coreference Models

Given the generally poorer performance of classification-based coreference models, a natural question is: can they be improved? To answer this question, we investigate whether these models can be improved by employing a different clustering algorithm and a different learning algorithm. There are reasons for our decision to focus on these two dimensions. First, as noted in the introduction, one of the weaknesses of these models is that it is not clear which clustering algorithm offers the best performance. Given this observation, we will examine whether we can improve these models by replacing Soon et al.'s (2001) "closest-first" linking regime with the "best-first" linking strategy, which has been shown to offer better performance for the mention-pair model on the MUC data sets (Ng & Cardie, 2002b). Second, as discussed at the end of Section 2, we may be able to achieve some of the advantage of ranking in classification-based models by employing a learning algorithm that optimizes for conditional probabilities instead of 0/1 decisions. Motivated by this observation, we will examine whether we can improve classification-based models by training them using MaxEnt, which employs a likelihood-based loss function. Note that MaxEnt is one of the most popular learning algorithms for training coreference models (see, for example, Morton, 2000; Kehler, Appelt, Taylor, & Simma, 2004; Ponzetto & Strube, 2006; Denis & Baldridge, 2008; Finkel & Manning, 2008; Ng, 2009).

---

21. Note that Haghighi and Klein did not report any CEAF scores in their paper.

To evaluate these two modifications, we apply them in isolation and in combination to the two classification-based models (i.e., the mention-pair model and the entity-mention model) when they are trained using three different feature sets (i.e., *Conventional*, *Lexical*, and *Combined*). We train the MaxEnt-based coreference models using YASMET[22], and follow Ng and Cardie's (2002b) implementation of the best-first clustering algorithm. Specifically, among the candidate antecedents or preceding clusters that are classified as coreferent with active mention $m_k$, best-first clustering links $m_k$ to the "most likely" one. For a MaxEnt model, a pair is classified as coreferent if and only if its classification value is above 0.5, and the most likely antecedent/preceding cluster for $m_k$ is the one that has the highest probability of coreference with $m_k$. For an SVM$^{light}$-trained model, a pair is classified as coreferent if and only if its classification value is above 0, and the most likely antecedent/preceding cluster for $m_k$ is the one that has the most positive classification value.

Table 5 presents B$^3$ and CEAF results of the two classification-based coreference models when they are trained using two learning algorithms (i.e., SVM and MaxEnt) and used in combination with two clustering algorithms (i.e., closest-first clustering and best-first clustering). To study how the choice of the clustering algorithm impacts performance, we should compare the results of closest-first clustering and best-first clustering in Table 5 for each combination of learning algorithm, feature set, coreference model, and scoring program. For instance, comparing rows 1 and 2 of Table 5 enables us to examine which of the two clustering algorithms is better for the mention-pair model when it is trained with the *Conventional* feature set and each of the two learners. Overall, we see a fairly consistent trend: best-first clustering yields results that are slightly worse than those obtained using closest-first clustering, regardless of the choice of the clustering algorithm, the learning algorithm, the feature set, and the scoring program. At first glance, these results seem contradictory to those by Ng and Cardie (2002b), who demonstrate the superiority of best-first clustering to closest-first clustering for coreference resolution. We speculate that the contradictory results can be attributed to two reasons. First, in our best-first clustering experiments, we still employed Soon et al.'s (2001) training instance selection method, where we created a positive training instance between an anaphoric mention and its closest antecedent/preceding cluster, unlike Ng and Cardie, who claim that "for the proposed best-first clustering to be successful, however, a different method for training instance selection would be needed." In particular, they propose to use the "most confident" antecedent, rather than the closest antecedent, to generate positive instances from an anaphoric mention. Second, Ng and Cardie demonstrate the success of best-first clustering on the MUC data sets, and it is possible that this success may not carry over to the ACE data sets. Additional experiments are needed to determine the reason, however.

---

22. See `http://www.fjoch.com/YASMET.html`. The reason why YASMET is chosen is that it provides the capability to *rank*, which allows us to compare the results of MaxEnt-trained classification models and ranking models. See the work of Ravichandran, Hovy, and Och (2003) for a discussion of the differences between the training of these two types of MaxEnt models.

| Coreference Model | SVM | | | MaxEnt | | |
|---|---|---|---|---|---|---|
| | R | P | F | R | P | F |

B[3] results using the *Conventional* feature set

| | Coreference Model | R | P | F | R | P | F |
|---|---|---|---|---|---|---|---|
| 1 | Mention-pair model (Closest first) | 46.2 | 72.0 | 56.2 | 59.6 | 55.3 | 57.3 |
| 2 | Mention-pair model (Best first) | 45.7 | 71.0 | 55.6 | 59.2 | 54.8 | 56.9 |
| 3 | Entity-mention model (Closest first) | 46.8 | 72.5 | 56.8 | 59.7 | 55.9 | **57.7** |
| 4 | Entity-mention model (Best first) | 46.3 | 72.1 | 56.3 | 59.3 | 55.1 | 57.1 |

B[3] results using the *Lexical* feature set

| | Coreference Model | R | P | F | R | P | F |
|---|---|---|---|---|---|---|---|
| 5 | Mention-pair model (Closest first) | 52.8 | 73.0 | **61.2** | 52.8 | 64.6 | 58.1 |
| 6 | Mention-pair model (Best first) | 52.1 | 72.1 | 60.5 | 52.1 | 64.2 | 57.5 |
| 7 | Entity-mention model (Closest first) | 52.8 | 73.6 | **61.2** | 52.8 | 64.6 | 58.2 |
| 8 | Entity-mention model (Best first) | 52.4 | 72.2 | 60.8 | 52.2 | 64.3 | 57.6 |

B[3] results using the *Combined* feature set

| | Coreference Model | R | P | F | R | P | F |
|---|---|---|---|---|---|---|---|
| 9 | Mention-pair model (Closest first) | 49.1 | 73.2 | 58.8 | 50.3 | 65.9 | 57.0 |
| 10 | Mention-pair model (Best first) | 48.7 | 72.8 | 58.3 | 49.3 | 65.2 | 56.1 |
| 11 | Entity-mention model (Closest first) | 49.5 | 73.2 | **59.1** | 50.5 | 66.1 | 57.3 |
| 12 | Entity-mention model (Best first) | 49.1 | 72.7 | 58.6 | 50.1 | 65.6 | 56.8 |

(a) B[3] results

CEAF results using the *Conventional* feature set

| | Coreference Model | R | P | F | R | P | F |
|---|---|---|---|---|---|---|---|
| 13 | Mention-pair model (Closest first) | 48.5 | 55.3 | 51.6 | 51.4 | 56.5 | 53.8 |
| 14 | Mention-pair model (Best first) | 48.1 | 54.9 | 51.2 | 51.1 | 56.1 | 53.4 |
| 15 | Entity-mention model (Closest first) | 49.5 | 55.8 | 52.5 | 51.4 | 56.7 | **53.9** |
| 16 | Entity-mention model (Best first) | 49.2 | 55.1 | 51.9 | 51.1 | 56.2 | 53.5 |

CEAF results using the *Lexical* feature set

| | Coreference Model | R | P | F | R | P | F |
|---|---|---|---|---|---|---|---|
| 17 | Mention-pair model (Closest first) | 54.6 | 61.2 | 57.7 | 53.5 | 56.8 | 55.1 |
| 18 | Mention-pair model (Best first) | 54.2 | 60.7 | 57.3 | 53.1 | 56.2 | 54.6 |
| 19 | Entity-mention model (Closest first) | 54.9 | 61.7 | **58.1** | 53.5 | 57.4 | 55.3 |
| 20 | Entity-mention model (Best first) | 54.5 | 61.1 | 57.6 | 53.1 | 56.9 | 54.9 |

CEAF results using the *Combined* feature set

| | Coreference Model | R | P | F | R | P | F |
|---|---|---|---|---|---|---|---|
| 21 | Mention-pair model (Closest first) | 53.8 | 60.0 | 56.7 | 54.9 | 56.3 | 55.5 |
| 22 | Mention-pair model (Best first) | 53.1 | 59.7 | 56.2 | 54.3 | 55.8 | 55.0 |
| 23 | Entity-mention model (Closest first) | 54.1 | 60.9 | **57.3** | 55.1 | 56.8 | 55.9 |
| 24 | Entity-mention model (Best first) | 53.7 | 60.3 | 56.8 | 54.5 | 56.2 | 55.3 |

(b) CEAF[3] results

Table 5: SVM vs. MaxEnt results for classification-based coreference models. These one-fold B[3] and CEAF scores are obtained by training coreference models using SVM and MaxEnt. The best F-measure achieved for each feature set/scoring program combination is boldfaced.

Next, to examine whether minimizing likelihood-based loss via MaxEnt training instead of SVM's classification loss would enable us to achieve some of the advantage of ranking (and hence leads to better performance), we compare the two columns of Table 5. As we can see, when the *Conventional* feature set is used, MaxEnt outperforms SVM, regardless of the choice of the clustering algorithm, the scoring program, and the coreference model. On the other hand, when the *Lexical* features or the *Combined* features are used, SVM outperforms MaxEnt consistently. Overall, these mixed results seem to suggest that whether MaxEnt offers better performance than SVM is to some extent dependent on the underlying feature set.

## 7.2 Performance of Maximum-Entropy-Based Ranking Models

Some prior work suggests that MaxEnt-based ranking may provide better gains than SVM-based ranking, since it can generate reliable confidence values and can dynamically adjust relative ranks according to baseline results (e.g., Ji, Rudin, & Grishman, 2006). To determine whether this is the case for coreference resolution, we conduct experiments in which we train the ranking-based coreference models using the ranker-learning algorithm in YAS-MET.

$B^3$ and CEAF results for the mention-ranking model and the cluster-ranking model when they are trained using MaxEnt in combination three different feature sets (i.e., *Conventional*, *Lexical*, and *Combined*) are shown in the "MaxEnt" column of Table 6. For comparison, we also show the corresponding results obtained via SVM-based ranking in the same table (see the "SVM" column). Comparing these two columns, we see mixed results: of the 24 experiments that involve ranking models, MaxEnt-based ranking outperforms SVM-based ranking in six of them. In other words, our results suggest that for the coreference task, SVM-based ranking is generally better than MaxEnt-based ranking.

## 7.3 Accuracy of Anaphoricity Determination

In Section 6.2, we saw that a joint ranking model always performs significantly better than its pipeline counterpart. In other words, joint modeling for coreference and anaphoricity improves coreference resolution. A natural question is: does joint modeling also improve anaphoricity determination?

To answer this question, we measure the accuracy of the anaphoricity information resulting from pipeline modeling and joint modeling. Recall that for pipeline modeling, we rely on the output of an anaphoricity classifier that is trained independently of the coreference system that uses the anaphoricity information (see Section 3.3). The accuracy of this classifier on the test set is shown under the "Acc" column in row 1 of Table 7. In addition, we show in the table its recall (R), precision (P), and F-measure (F) on identifying *anaphoric* mentions. As we can see, the classifier achieves an accuracy of 81.1 and a F-measure score of 83.8.

On the other hand, for joint modeling, we can compute the accuracy of anaphoricity determination from the output of a joint coreference model. Specifically, given the output of a joint model, we can determine which mentions are resolved to a preceding antecedent and which are not. Assuming that a mention that is resolved is anaphoric and one that is not resolved is non-anaphoric, we can compute the accuracy of anaphoricity determination

| Coreference Model | SVM | | | MaxEnt | | |
|---|---|---|---|---|---|---|
| | R | P | F | R | P | F |

B[3] results using the *Conventional* feature set

| | | R | P | F | R | P | F |
|---|---|---|---|---|---|---|---|
| 1 | Mention-ranking model (Pipeline) | 46.7 | 71.5 | 56.5 | 58.7 | 59.1 | 58.8 |
| 2 | Mention-ranking model (Joint) | 47.6 | 74.8 | 58.2 | 59.1 | 59.3 | 59.2 |
| 3 | Cluster-ranking model (Pipeline) | 53.6 | 59.5 | 56.4 | 51.7 | 69.9 | 59.3 |
| 4 | Cluster-ranking model (Joint) | 52.2 | 73.8 | **61.2** | 52.1 | 70.6 | 60.0 |

B[3] results using the *Lexical* feature set

| | | R | P | F | R | P | F |
|---|---|---|---|---|---|---|---|
| 5 | Mention-ranking model (Pipeline) | 53.8 | 68.3 | 60.1 | 56.6 | 61.1 | 58.8 |
| 6 | Mention-ranking model (Joint) | 54.6 | 72.8 | **62.4** | 56.3 | 64.4 | 60.1 |
| 7 | Cluster-ranking model (Pipeline) | 51.7 | 68.2 | 58.8 | 48.4 | 66.6 | 56.1 |
| 8 | Cluster-ranking model (Joint) | 52.9 | 73.4 | 61.5 | 48.0 | 72.9 | 57.8 |

B[3] results using the *Combined* feature set

| | | R | P | F | R | P | F |
|---|---|---|---|---|---|---|---|
| 9 | Mention-ranking model (Pipeline) | 49.8 | 72.6 | 59.1 | 51.4 | 68.3 | 58.7 |
| 10 | Mention-ranking model (Joint) | 50.5 | 77.6 | 61.2 | 52.5 | 70.3 | 60.1 |
| 11 | Cluster-ranking model (Pipeline) | 53.8 | 71.2 | 61.3 | 54.1 | 67.5 | 60.1 |
| 12 | Cluster-ranking model (Joint) | 54.4 | 74.8 | **62.8** | 54.5 | 68.3 | 60.6 |

(a) B[3] results

CEAF results using the *Conventional* feature set

| | | R | P | F | R | P | F |
|---|---|---|---|---|---|---|---|
| 13 | Mention-ranking model (Pipeline) | 49.4 | 55.7 | 52.4 | 51.5 | 56.6 | 53.9 |
| 14 | Mention-ranking model (Joint) | 50.5 | 56.3 | 53.2 | 51.8 | 56.9 | 54.2 |
| 15 | Cluster-ranking model (Pipeline) | 53.6 | 59.5 | 56.4 | 53.1 | 58.7 | 55.8 |
| 16 | Cluster-ranking model (Joint) | 55.2 | 61.6 | **58.2** | 53.2 | 59.3 | 56.1 |

CEAF results using the *Lexical* feature set

| | | R | P | F | R | P | F |
|---|---|---|---|---|---|---|---|
| 17 | Mention-ranking model (Pipeline) | 54.7 | 59.8 | 57.1 | 55.7 | 56.6 | 56.1 |
| 18 | Mention-ranking model (Joint) | 56.9 | 63.3 | **59.9** | 55.4 | 60.7 | 57.9 |
| 19 | Cluster-ranking model (Pipeline) | 52.7 | 58.4 | 55.4 | 50.9 | 52.3 | 51.5 |
| 20 | Cluster-ranking model (Joint) | 55.0 | 61.4 | 58.1 | 50.5 | 56.6 | 53.3 |

CEAF results the *Combined* feature set

| | | R | P | F | R | P | F |
|---|---|---|---|---|---|---|---|
| 21 | Mention-ranking model (Pipeline) | 53.7 | 58.8 | 56.1 | 53.4 | 58.7 | 55.9 |
| 22 | Mention-ranking model (Joint) | 54.9 | 61.7 | 58.1 | 54.9 | 56.3 | 55.5 |
| 23 | Cluster-ranking model (Pipeline) | 55.1 | 60.1 | 57.4 | 56.1 | 60.0 | 57.9 |
| 24 | Cluster-ranking model (Joint) | 58.4 | 65.1 | **61.6** | 56.7 | 62.5 | 59.5 |

(b) CEAF results

Table 6: SVM vs. MaxEnt results for ranking-basd coreference models. These one-fold B[3] and CEAF scores are obtained by training coreference models using SVM and MaxEnt. The best F-measure achieved for each feature set/scoring program combination is boldfaced.

| | Source of Anaphoricity Information | Acc | R | P | F |
|---|---|---|---|---|---|
| 1 | Anaphoricity Classifier | 81.1 | 87.6 | 80.3 | 83.8 |
| 2 | Mention-ranking (*Conventional*) | 78.7 | 83.1 | 79.3 | 81.2 |
| 3 | Cluster-ranking (*Conventional*) | 81.9 | 87.8 | 80.4 | 83.9 |
| 4 | Mention-ranking (*Lexical*) | 84.2 | 88.3 | 82.1 | 85.1 |
| 5 | Cluster-ranking (*Lexical*) | 83.1 | 87.9 | 81.8 | 84.7 |
| 6 | Mention-ranking (*Combined*) | 79.1 | 84.3 | 79.1 | 81.6 |
| 7 | Cluster-ranking (*Combined*) | 83.1 | 87.4 | 82.1 | 84.6 |

Table 7: Anaphoricity determination results.

as well as the precision, recall, and F-measure on identifying anaphoric mentions. Since all these performance numbers are derived from the output of a joint model, we can compute them for each of the two joint ranking models (i.e., the mention-ranking model and the cluster-ranking model) when used in combination with each of the three coreference feature sets (i.e., *Conventional*, *Lexical*, and *Combined*). This results in six sets of performance numbers, which are shown in rows 2–7 of Table 4. As we can see, the accuracies range from 78.7 to 84.2, and the F-measure scores range from 81.2 to 85.1.

In comparison to the results of the anaphoricity classifier shown in row 1, we can see that joint modeling improves the performance of anaphoricity determination except for two cases, namely, mention-ranking/*Conventional* and mention-ranking/*Combined*. In other words, in these two cases, joint modeling benefits coreference resolution but not anaphoricity determination. While it seems counter-intuitive that one can achieve better coreference performance with a lower accuracy on determining anaphoricity, it should not be difficult to see the reason: the joint model is trained to maximize the pairwise ranking accuracy, which presumably correlates with coreference performance, whereas the anaphoricity classifier is trained to maximize the accuracy of determining the anaphoricity of a mention, which may not *always* have any correlation with coreference performance. In other words, improvements in anaphoricity accuracy generally but not necessarily imply corresponding improvements in clustering-level coreference accuracy.

Finally, it is important to bear in mind that the conclusions we have drawn regarding pipeline and joint modeling are based on the results of an anaphoricity classifier trained on 26 features. It is possible that different conclusions could be drawn if we trained the anaphoricity classifier on a different set of features. Therefore, an interesting future direction would be to improve the anaphoricity classifier by employing additional features, such as those proposed by Uryupina (2003). We may also be able to derive sophisticated features by harnessing recent advances in lexical semantics research, specifically by using methods for phrase clustering (e.g., Lin & Wu, 2009), lexical chain discovery (e.g., Morris & Hirst, 1991), and paraphrase discovery (see the survey papers by Androutsopoulos & Malakasiotis, 2010; Madnani & Dorr, 2010).

## 7.4 Joint Inference Versus Joint Learning for the Mention-Pair Model

As mentioned at the end of Section 4.2, joint modeling for anaphoricity determination and coreference resolution is fundamentally different from joint inference for these two tasks. Recall that in joint inference using ILP, an anaphoricity classifier and a coreference classifier are trained independently of each other, and then ILP is applied as a postprocessing step to jointly infer anaphoricity and coreference decisions so that they are consistent with each other (e.g., Denis & Baldridge, 2007a). In this subsection, we investigate how joint learning compares with joint inference for anaphoricity determination and coreference resolution.

Let us begin with an overview of the ILP approach proposed by Denis and Baldridge (2007a) for joint inference for anaphoricity determination and coreference resolution. The ILP approach is motivated by the observation that the output of an anaphoricity model and that of a coreference model for a given document have to satisfy certain constraints. For instance, if the coreference model determines that a mention $m_k$ is not coreferent with any other mentions in the associated text, then the anaphoricity model should determine that $m_k$ is non-anaphoric. In practice, however, since the two models are trained independently of each other, this and other constraints cannot be enforced.

Denis and Baldridge (2007a) provide an ILP framework for *jointly* determining anaphoricity and coreference decisions for a given set of mentions based on the probabilities provided by the anaphoricity model $P_A$ and the mention-pair coreference model $P_C$, such that the resulting joint decisions satisfy the desired constraints while respecting as much as possible the probabilistic decisions made by the independently-trained $P_A$ and $P_C$. Specifically, an ILP program is composed of an objective function to be optimized subject to a set of linear constraints, and is created for each test text $D$ as follows. Let $M$ be the set of mentions in $D$, and $P$ be the set of mention pairs formed from $M$ (i.e., $P = \{(m_j, m_k) \mid m_j, m_k \in M, j < k\}$). Each ILP program has a set of indicator *variables*. In our case, we have one binary-valued variable for each anaphoricity decision and coreference decision to be made by an ILP solver. Following Denis and Baldridge's notation, we use $y_k$ to denote the anaphoricity decision for mention $m_k$, and $x_{\langle j,k \rangle}$ to denote the coreference decision involving mentions $m_j$ and $m_k$. In addition, each variable is associated with an *assignment* cost. Specifically, let $c^C_{\langle j,k \rangle} = -\log(P_C(m_j, m_k))$ be the cost of setting $x_{\langle j,k \rangle}$ to 1, and $\bar{c}^C_{\langle j,k \rangle} = -\log(1 - P_C(m_j, m_k))$ be the complementary cost of setting $x_{\langle j,k \rangle}$ to 0. We can similarly define the cost associated with each $y_k$, letting $c^A_k = -\log(P_A(m_k))$ be the cost of setting $y_k$ to 1, and $\bar{c}^A_k = -\log(1 - P_A(m_k))$ be the complementary cost of setting $y_k$ to 0. Given these costs, we aim to optimize the following objective function:

$$\min \sum_{(m_j, m_k) \in P} c^C_{\langle j,k \rangle} \cdot x_{\langle j,k \rangle} + \bar{c}^C_{\langle j,k \rangle} \cdot (1 - x_{\langle j,k \rangle}) + \sum_{m_k \in M} c^A_k \cdot y_k + \bar{c}^A_k \cdot (1 - y_k)$$

subject to a set of manually-specified *linear* constraints. Denis and Baldridge specify four types of constraints: (1) each indicator variable can take on a value of 0 or 1; (2) if $m_j$ and $m_k$ are coreferent ($x_{\langle j,k \rangle}$=1), then $m_k$ is anaphoric ($y_k$=1); (3) if $m_k$ is anaphoric ($y_k$=1), then it must be coreferent with some preceding mention $m_j$; and (4) if $m_k$ is non-anaphoric, then it cannot be coreferent with any mention.

Two points deserve mention. First, we are *minimizing* the objective function, since each assignment cost is expressed as a negative logarithm value. Second, since transitivity is not guaranteed by the above constraints[23], we use the closest-link clustering algorithm to put any two mentions that are posited as coreferent into the same cluster. Note that the best-link clustering strategy is not applicable here, since a binary decision is assigned to each pair of mentions by the ILP solver. We use *lp_solve*[24], a publicly-available ILP solver, to solve this program.

$B^3$ and CEAF results of performing joint inference on the outputs of the anaphoricity model and the mention-pair model using ILP are shown in the "Joint Inference" column of Tables 8a and 8b, respectively, where the rows correspond to results obtained by training the coreference models on different feature sets. Since one of our goals is to compare joint inference and joint learning, we also show in the "Joint Learning" column the results of the joint mention-ranking model, where anaphoricity determination and coreference resolution are learned in a joint fashion. Note that the reason for using the mention-ranking model (rather than the cluster-ranking model) as our joint model here is that we want to ensure a fair comparison of joint learning and joint inference as much as possible: had we chosen the cluster-ranking model as our joint model, the difference between the joint learning results and the joint inference results could have been caused by the increased expressiveness of the cluster-ranking model. Finally, to better understand whether the mention-pair model benefits from joint inference using ILP, we show in the "No Inference" column the relevant mention-pair model results from Table 4, where the output of the model is not postprocessed with any inference mechanism.

From Table 8, we can see that the joint learning results are substantially better than the joint inference results, except for one case (*Conventional*/CEAF), where the two achieve comparable performance. Previous work by Roth (2002) and Roth and Yih (2004) has suggested that it is often more effective to learn simple local models and use complicated integration strategies to make sure constraints on the output are satisfied than to learn models that satisfy the constraints directly. Our results imply that this is not true for the coreference task.

Comparing the joint inference and "No Inference" results in Table 8, we can see that the mention-pair model does not benefit from the application of ILP. In fact, its performance deteriorates when ILP is used. These results are inconsistent with those reported by Denis and Baldridge (2007a), who show that joint inference using ILP can improve the mention-pair model. We speculate that the inconsistency accrues from the fact that Denis and Baldridge evaluate the ILP approach on *true* mentions (i.e., gold-standard mentions), while we evaluate it on system mentions. Additional experiments are needed to determine the reason, however.

## 7.5 Data Source Adaptability

One may argue that since we train and test a model on documents from the same data source (i.e., the model trained on the documents from BC is tested on the documents from

---

23. Finkel and Manning (2008) show how to formulate linear constraints so that the ILP solver outputs coreference decisions that satisfy transitivity. However, since the number of additional constraints needed to guarantee transitivity grows cubically with the number of mentions and previous work shows that having these additional constraints do not yield substantial performance improvements when applied to system mentions (Ng, 2009), we decided not to employ them in our experiments.

24. Available from `http://lpsolve.sourceforge.net/`

|   | Feature Set | Joint Learning | | | Joint Inference | | | No Inference | | |
|---|---|---|---|---|---|---|---|---|---|---|
|   |   | R | P | F | R | P | F | R | P | F |
| 1 | *Conventional* | 47.6 | 74.8 | 58.2 | 58.2 | 55.9 | 57.0 | 59.6 | 55.3 | 57.3 |
| 2 | *Lexical* | 54.6 | 72.8 | 62.4 | 49.1 | 70.1 | 57.8 | 52.8 | 64.6 | 58.1 |
| 3 | *Combined* | 50.5 | 77.6 | 61.2 | 53.2 | 56.9 | 54.9 | 50.3 | 65.9 | 57.0 |

(a) B³ results

|   | Feature Set | Joint Learning | | | Joint Inference | | | No Inference | | |
|---|---|---|---|---|---|---|---|---|---|---|
|   |   | R | P | F | R | P | F | R | P | F |
| 1 | *Conventional* | 50.5 | 56.3 | 53.2 | 49.7 | 57.5 | 53.3 | 51.4 | 56.5 | 53.8 |
| 2 | *Lexical* | 56.9 | 63.3 | 59.9 | 50.6 | 58.6 | 54.3 | 53.5 | 56.8 | 55.1 |
| 3 | *Combined* | 54.9 | 61.7 | 58.1 | 53.2 | 56.9 | 54.9 | 54.9 | 56.3 | 55.5 |

(b) CEAF results

Table 8: Joint learning vs. joint inference results. The joint modeling results are obtained using the mention-ranking model. The joint inference results are obtained by applying ILP to the anaphoricity classifier and the mention-pair model. The "no inference" results are those produced by the mention-pair model. All coreference models are trained using MaxEnt.

BC, for example), it should not be surprising that lexicalization helps, since word pairs in a training set are more likely to be found in a test set if the training and test texts are from the same data source. To examine whether models that employ the *Lexical* features will suffer if they are trained and tested on different data sources, we perform a set of *data source adaptability* experiments, where we apply a coreference model that is trained with the *Lexical* features on documents from one data source to documents from *all* data sources. Here, we show the results obtained using the mention-ranking model, primarily because it yielded the best performance with the *Lexical* features among the learning-based coreference models. For comparison, we also show the data source adaptability results obtained using the mention-ranking model that is trained with the (non-lexical) *Conventional* feature set.

The B³ and CEAF F-measure scores of these experiments are shown in Tables 9a and 9b, where the left half and the right half of the table contain the lexicalized mention-ranking model results and the unlexicalized mention-ranking model results, respectively. Each row corresponds to a data source on which a model is trained, except for the last two rows, which we will explain shortly. Each column corresponds to a test set from a particular data source.

To answer the question of whether the performance of a coreference model that employs the *Lexical* features will deteriorate if they are trained and tested on different data sources, we can look at the diagonal entries in the left half of Tables 9a and 9b, which contain the results obtained when the lexicalized mention-ranking model is trained and tested on documents from the same source. If the model indeed performs worse when it is trained and tested on documents from different sources, then a diagonal entry should contain the highest score among the entries in the same column. As we can see from the left half of the two tables, this is to a large extent correct: four of the six diagonal entries contain the highest scores in their respective columns according to both scoring programs. This provides

| Test |  Lexical features | | | | | | Conventional features | | | | | |
|---|---|---|---|---|---|---|---|---|---|---|---|---|
| Train | BC | BN | CTS | NW | UN | WL | BC | BN | CTS | NW | UN | WL |
| BC | 56.5 | 61.0 | 58.7 | 64.2 | **57.2** | 66.6 | 52.1 | 55.9 | 55.1 | 59.1 | 52.8 | 63.8 |
| BN | **57.6** | **63.5** | 60.7 | 63.5 | 55.4 | 67.0 | 51.8 | **59.7** | 58.4 | 58.7 | 52.5 | 62.8 |
| CTS | 55.5 | 61.1 | **62.7** | 61.9 | 54.9 | 65.9 | 51.8 | 58.4 | **59.5** | **59.2** | **53.6** | 64.2 |
| NW | 55.6 | 62.1 | 56.7 | **65.4** | 56.4 | 59.2 | 51.5 | 55.3 | 55.7 | 58.5 | 52.1 | 60.3 |
| UN | 56.4 | 62.8 | 60.0 | 62.9 | 56.3 | 67.3 | **52.7** | 57.3 | 59.2 | 59.0 | 53.2 | **64.2** |
| WL | 56.4 | 62.8 | 58.5 | 63.4 | 55.9 | **68.7** | 51.5 | 56.5 | 55.1 | 59.0 | 52.7 | 64.0 |
| Max−Min | 2.1 | 2.5 | 6.0 | 3.5 | 2.3 | 9.5 | 1.2 | 4.4 | 4.4 | 0.7 | 1.5 | 3.9 |
| Std. Dev. | 0.76 | 1.01 | 2.07 | 1.18 | 0.81 | 3.36 | 0.45 | 1.64 | 2.09 | 0.26 | 0.53 | 1.52 |

(a) B³ results

| Test |  Lexical features | | | | | | Conventional features | | | | | |
|---|---|---|---|---|---|---|---|---|---|---|---|---|
| Train | BC | BN | CTS | NW | UN | WL | BC | BN | CTS | NW | UN | WL |
| BC | 52.0 | 57.2 | 55.5 | 61.5 | **56.5** | 65.7 | **46.0** | 49.3 | 52.7 | 53.1 | 49.1 | 60.8 |
| BN | **53.8** | **61.3** | 58.8 | 60.5 | 54.1 | 66.0 | 45.9 | **54.8** | 55.2 | 52.9 | 48.0 | 60.1 |
| CTS | 51.9 | 58.6 | **62.0** | 58.4 | 53.7 | 64.8 | 44.7 | 52.3 | **56.5** | **53.8** | 50.0 | 61.0 |
| NW | 50.5 | 58.9 | 53.3 | **62.7** | 54.6 | 54.5 | 45.5 | 49.2 | 52.6 | 52.7 | 48.4 | 56.5 |
| UN | 52.5 | 60.3 | 58.3 | 59.3 | 55.7 | 67.0 | **46.0** | 51.3 | 56.2 | 52.8 | 48.8 | 60.8 |
| WL | 53.4 | 61.1 | 55.7 | 59.8 | 55.0 | **67.1** | 45.7 | 50.4 | 51.4 | 52.9 | **50.1** | **61.1** |
| Max−Min | 3.3 | 4.1 | 8.7 | 4.3 | 2.8 | 12.7 | 1.3 | 5.6 | 5.1 | 1.1 | 2.1 | 4.6 |
| Std. Dev. | 1.18 | 1.60 | 3.07 | 1.55 | 1.04 | 4.82 | 0.50 | 2.11 | 2.14 | 0.40 | 0.85 | 1.77 |

(b) CEAF results

Table 9: Results for data source adaptability. Each row shows the results obtained by training the mention ranking model on the data set shown on the first column of the row, and each column corresponds to the test set from a particular data source. The best result obtained on each test set for each of the two coreference models is boldfaced.

suggestive evidence that the answer to our question is affirmative. Nevertheless, if we look at right half of the two tables, where we show the results obtained using the unlexicalized mention-ranking model, we see a similar, but perhaps weaker, trend: according to CEAF, four of the six diagonal entries contain the highest scores in their respective columns, and according to B³, two of the six diagonal entries exhibit this trend. Hence, the fact that a model performs worse when it is trained and tested on different data sources cannot be attributed solely to lexicalization.

Perhaps a more informative question is: do lexicalized models trained on different data sources exhibit more varied performance on a given test set (composed of documents from the same source) than unlexicalized models trained on different data sources? An affirmative answer to this question will provide empirical support for the hypothesis that a lexicalized model fits the data on which it is trained more than its unlexicalized counterpart. To answer this question, we compute for each column and each of the two models (1) the difference between the highest and lowest scores (see the "Max−Min" row), and (2) the standard deviation of the six scores in the corresponding column (see the "Std. Dev." row). If we compare the corresponding columns of the two coreference models, we can see that except for BN, the lexicalized model does exhibit more varied performance on a given test set than the unlexicalized model according to both scoring programs, regardless of whether we are measuring the variation using Max−Min or standard deviation.

## 7.6 Feature Analysis

In this subsection, we analyze the effects of the linguistic features on the performance of the coreference models. Given the large number of models trained on each of the three feature sets, it is not feasible for us to analyze the features for each model and each feature set. Since the cluster-ranking model, when used with the *Combined* feature set, yields the best performance, we will analyze its features. In addition, since the *Lexical* features have yielded good performance for the mention-ranking model, it would be informative to see which *Lexical* features have the greatest contribution to its performance. As a result, we will perform feature analysis on these two model/feature set combinations.

Although we have identified two particular model/feature set combinations, we actually have a total of 12 model/feature set combinations: recall that except for row 1, each row in Table 4 shows the aggregated result for the six data sets, where we trained one model for each data set. In other words, for each of the two combinations we selected above, we have six learned models. To reduce the number of models we need to analyze and yet maximize the insights we can gain, we choose to analyze the models that were trained on data sets from two fairly different domains: Newswire (NW) and Broadcast News (BN).

The next question is: how can we analyze the features? We apply the *backward elimination* feature selection algorithm (see the survey paper by Blum & Langley, 1997), which starts with the full feature set and removes in each iteration the feature whose removal yields the best system performance. Despite its greedy nature, this algorithm runs in time quadratic to the number of features, making it computationally expensive to run on our feature sets. To reduce computational cost, we divide our features into *feature types* and apply backward elimination to eliminate one feature type per iteration.

The features are grouped as follows. For the *Lexical* feature set, we divide the features into five types: (1) unseen features, (2) lexical features, (3) semi-lexical features, (4) DIS-TANCE, and (5) ALIAS. In other words, the division corresponds roughly to the one described in Section 5.1, except that we put the two "conventional" features into two different groups, since linguistically one is a positional feature and the other is a semantic feature. For the *Combined* feature set, we divide the features into seven groups, the first four of which are identical to those in the division of the *Lexical* features above. For the remaining features, we divide them into STRING-MATCHING features, which comprise features 11–18 in Table 1; GRAMMATICAL features, which comprise features 1–7, 9–10, 19–29, 33–36, and 38–39; and SEMANTIC features, which comprise features 8, 30, and 31. Note that ALIAS, the only semantic feature in the *Lexical* feature set, is combined with other semantic features in the *Conventional* feature set to form the SEMANTIC feature type.

Results are shown in Tables 10–13. Specifically, Tables 10a and 10b show the $B^3$ and CEAF F-measure scores of the feature analysis experiments involving the mention-ranking model, using the *Lexical* feature set on the NW data set. In each table, the first row shows how the system would perform if each class of features were removed. We remove the least

| Lexical | Semi-lexical | Distance | Alias | Unseen |
|---------|--------------|----------|-------|--------|
| 60.6 | 62.4 | 63.3 | 64.6 | 65.2 |
| 60.1 | 53.1 | 60.1 | 63.5 | |
| 59.2 | 59.7 | 61.8 | | |
| 60.7 | 60.9 | | | |

(a) $B^3$ results

| Semi-lexical | Lexical | Distance | Alias | Unseen |
|--------------|---------|----------|-------|--------|
| 58.4 | 56.0 | 60.7 | 61.7 | 62.1 |
| 44.7 | 55.0 | 55.3 | 59.3 | |
| 53.4 | 53.5 | 58.7 | | |
| 54.7 | 55.6 | | | |

(b) CEAF results

Table 10: Feature analysis results (in terms of F-measure scores) for the mention-ranking model using the *Lexical* features on the NW data set. When all feature types are used to train the model, the $B^3$ and CEAF F-measure scores are 65.4 and 62.7, respectively.

important feature class (i.e., the feature class whose removal yields the best performance), and the next row shows how the adjusted system would perform without each remaining class. According to both scoring programs, removing the UNSEEN features yields the least drop to performance (note from the caption that with the full feature set, the $B^3$ score is 65.4 and the CEAF score is 62.7). In fact, the two scorers agree that the LEXICAL and SEMI-LEXICAL features are more important than the UNSEEN, ALIAS, and DISTANCE features. Nevertheless, these results suggest that all five feature types are important, since the best performance is achieved using the full feature set.

Tables 11a and 11b show the $B^3$ and CEAF F-measure scores of the feature analysis experiments involving the cluster-ranking model, using the *Combined* feature set on the NW data set. Recall that in the *Combined* feature set, we have seven types of features. As we can see, the two scorers agree completely on the order in which the features should be removed. In particular, the most important features are the LEXICAL and SEMI-LEXICAL features, whereas the least important features are those that are not present in the *Lexical* feature set, namely, the GRAMMATICAL, STRING-MATCHING, and SEMANTIC features. This suggests that the lexical features are in general more important than the non-lexical features when they are used in combination. This is somewhat surprising, as the non-lexical features are the commonly-used features for coreference resolution, whereas *Lexical* features are comparatively much less investigated by coreference researchers. Nevertheless, unlike what we saw in Table 10, where all feature types appear to be relevant, in Table 11a, we see that

| Lexical | Semi-lexical | Unseen | Distance | Semantic | String match | Grammatical |
|---------|--------------|--------|----------|----------|--------------|-------------|
| 54.7 | 59.7 | 58.5 | 57.6 | 59.7 | 59.6 | 60.4 |
| 56.6 | 60.4 | 58.8 | 58.6 | 60.2 | 61.1 | |
| 58.9 | 61.1 | 58.9 | 59.9 | 62.4 | | |
| 58.3 | 62.3 | 57.5 | 63.9 | | | |
| 63.2 | 61.9 | 63.8 | | | | |
| 63.2 | 67.2 | | | | | |

(a) $B^3$ results

| Lexical | Semi-lexical | Unseen | Distance | Semantic | String match | Grammatical |
|---------|--------------|--------|----------|----------|--------------|-------------|
| 50.2 | 53.3 | 56.9 | 55.6 | 57.7 | 58.4 | 58.7 |
| 50.6 | 54.3 | 57.6 | 56.9 | 58.5 | 60.6 | |
| 51.4 | 57.1 | 58.2 | 58.8 | 60.1 | | |
| 51.0 | 57.2 | 56.7 | 60.0 | | | |
| 57.9 | 55.3 | 60.0 | | | | |
| 57.9 | 61.9 | | | | | |

(b) CEAF results

Table 11: Feature analysis results (in terms of F-measure) for the cluster-ranking model using the *Combined* features on the NW data set. When all feature types are used to train the model, the $B^3$ and CEAF F-measure scores are 64.6 and 62.3, respectively.

the best $B^3$ F-measure score is 67.2, which is achieved using only the LEXICAL features. This represents a 2.6% absolute gain in F-measure over the model trained on all seven feature types, suggesting a learning-based coreference model could be improved via feature selection.

Next, we investigate whether similar trends can be observed when the models are trained on a different source: Broadcast News. Specifically, we show in Tables 12a and 12b the $B^3$ and CEAF F-measure scores of the feature analysis experiments involving the mention-ranking model, using the *Lexical* feature set on the BN data set. As in Table 11, we see that the two scorers agree completely on the order in which the features should be removed. In fact, similar to what we observed in Table 10 (on the NW data set), both scorers determine that the LEXICAL and SEMI-LEXICAL features are the most important, whereas the DISTANCE and ALIAS features are the least important, although all five feature types appear to be relevant according to both scorers.

Finally, we show in Tables 13a and 13b the $B^3$ and CEAF F-measure scores of the feature analysis experiments involving the cluster-ranking model, using the *Combined* feature set

| Lexical | Semi-lexical | Distance | Unseen | Alias |
|---------|--------------|----------|--------|-------|
| 53.4 | 62.7 | 60.9 | 62.9 | 63.4 |
| 53.4 | 62.3 | 59.9 | 62.9 | |
| 52.3 | 61.2 | 61.7 | | |
| 52.9 | 61.6 | | | |

(a) B$^3$ results

| Lexical | Semi-lexical | Distance | Unseen | Alias |
|---------|--------------|----------|--------|-------|
| 47.1 | 59.6 | 58.6 | 59.9 | 61.2 |
| 47.1 | 59.1 | 57.8 | 60.0 | |
| 44.9 | 57.4 | 58.0 | | |
| 45.5 | 57.5 | | | |

(b) CEAF results

Table 12: Feature analysis results (in terms of F-measure) for the mention-ranking model using the *Lexical* features on the BN data set. When all feature types are used to train the model, the B$^3$ and CEAF F-measure scores are 63.5 and 61.3, respectively.

on the BN data set. As in Tables 11 and 12, the two scorers agree completely on the order in which the features should be removed. As far as feature contribution is concerned, these two tables resemble Tables 11a and 11b: in both cases, the LEXICAL, SEMI-LEXICAL, and UNSEEN features are the most important; the STRING-MATCHING and GRAMMATICAL features are the least important; and the SEMANTIC and DISTANCE features are in the middle. In this case, however, all seven feature types seem to be relevant, as the best performance is achieved using the full feature set according to both scorers. Perhaps most interestingly, the numbers in each column are generally increasing as we move down the column. This means that a feature type becomes progressively less useful as we remove more and more feature types. This also suggests that the interactions between different feature types are non-trivial and that a feature type may be useful only in the presence of another feature type.

In summary, results on two data sets (NW and BN) and two scoring programs demonstrate that (1) in general all feature types are crucial to overall performance, and (2) the little-investigated *Lexical* features contribute more to overall performance than the commonly-used *Conventional* features.

## 7.7 Resolution Performance

To gain additional insights into our results, we analyze the behavior of the coreference models for different types of anaphoric expressions when they are trained with different feature sets. Specifically, we partition the mentions into different *resolution classes*. While

| Lexical | Semi-lexical | Unseen | Semantic | Distance | String match | Grammatical |
|---------|--------------|--------|----------|----------|--------------|-------------|
| 54.7 | 52.7 | 52.3 | 51.2 | 51.8 | 53.9 | 54.8 |
| 55.2 | 54.2 | 53.9 | 54.3 | 53.3 | 55.6 | |
| 55.4 | 56.9 | 56.6 | 60.6 | 56.6 | | |
| 55.5 | 60.4 | 59.5 | 61.3 | | | |
| 56.9 | 57.4 | 61.3 | | | | |
| 56.9 | 60.6 | | | | | |

(a) $B^3$ results

| Lexical | Semi-lexical | Unseen | Semantic | Distance | String match | Grammatical |
|---------|--------------|--------|----------|----------|--------------|-------------|
| 44.4 | 45.7 | 45.2 | 44.3 | 47.1 | 46.4 | 52.1 |
| 44.3 | 51.3 | 50.7 | 51.0 | 50.0 | 51.8 | |
| 45.5 | 52.9 | 54.9 | 54.0 | 58.5 | | |
| 46.3 | 56.7 | 57.0 | 57.7 | | | |
| 48.5 | 50.5 | 57.1 | | | | |
| 48.6 | 55.4 | | | | | |

(b) CEAF results

Table 13: Feature analysis results (in terms of F-measure) for the cluster-ranking model using the *Combined* features on the BN data set. When all feature types are used to train the model, the $B^3$ and CEAF F-measure scores are 63.6 and 61.3, respectively.

previous work has focused mainly on three rather coarse-grained resolution classes (namely, pronouns, proper nouns, and common nouns), we follow Stoyanov et al. (2009) and subdivide each class into three fine-grained classes. It is worth mentioning that none of Stoyanov et al.'s classes corresponds to non-anaphoric expressions. Since we believe that non-anaphoric expressions play an important role in the analysis of the performance of a coreference model, we propose three additional classes that correspond to non-anaphoric pronouns, non-anaphoric proper nouns, and non-anaphoric common nouns. Finally, there are certain types of anaphoric pronouns (e.g., wh-pronouns) that do not fall into any of Stoyanov et al.'s pronoun categories. To fill this gap, we create another category that serves as the default category for any anaphoric pronouns not covered by Stoyanov et al.'s classes. This results in 13 resolution classes, which are discussed below in detail.

**Proper nouns.** Four classes are defined for proper nouns. (1) **e**: a proper noun is assigned to this *exact string match* class if there is a preceding mention such that the two are coreferent and are the same string; (2) **p**: a proper noun is assigned to this *partial string match* class if there is a preceding mention such that the two are coreferent and have some content words in common; (3) **n**: a proper noun is assigned to this *no string match* class if there is no preceding mention such that the two are coreferent and have some content words in common; and (4) **na**: a proper noun is assigned to this *non-anaphor* class if it is not coreferent with any preceding mention.

**Common nouns.** Four analogous resolution classes are defined for mentions whose head is a common noun: (5) **e**; (6) **p**; (7) **n**; and (8) **na**.

**Pronouns.** We have three pronoun classes. (9) **1+2**: 1st and 2nd person pronouns; (10) **G3**: gendered 3rd person pronouns (e.g., *she*); (11) **U3**: ungendered 3rd person pronouns; (12) **oa**: any anaphoric pronouns that do not belong to (9), (10), and (11); and (13) **na**: non-anaphoric pronouns.

Next, we score each resolution class. Unlike Stoyanov et al. (2009), who use a modified version of the MUC scorer, we employ $B^3$. The reasons are that the MUC scorer (1) does not reward singleton clusters, and (2) can inflate a system's performance when the clusters are overly large. To compute the score for class C, we process the mentions in a test text in a left-to-right manner. For each mention encountered, we check whether it belongs to C. If so, we use our coreference model to decide how to resolve it. Otherwise, we use an oracle to make the correct resolution decision[25] (so that in the end all the mistakes can be attributed to the incorrect resolution of the mentions in C, thus allowing us to directly measure its impact on overall performance). After all the test documents are processed, we compute the $B^3$ F-measure score on only the mentions that belong to C.

Performance of each resolution class, when aggregated over the test sets of the six data sources in the same way as before, are shown in Table 14, which provides a nice diagnosis of the strengths and weaknesses of each coreference model when used in combination with each feature set. We also show in the table the percentage of mentions belonging to each class below the name of each class, and abbreviate the name of each model as follows: HM corresponds to the head match baseline, whereas MP, EM, MR, and CR denote the mention-pair model, the entity-mention model, the mention-ranking model, and the cluster-ranking model, respectively. Each ranking model has two versions, the pipeline version (denoted by P) and the joint version (denoted by J).

A few points deserve mention. Recall from Table 4 that when the *Conventional* features are used, the joint mention-ranking model performs better than the mention-pair model and the entity-mention model. Comparing row 5 with rows 2 and 3 of Table 14, we can see that the improvements can be attributed primarily to its better handling of one proper

---

25. If the oracle determines that a mention is anaphoric and that its antecedents are not in the same cluster (because our model has previously made a mistake), we employ the following heuristic to select which antecedent to resolve the mention to: we try to resolve it to the closest preceding antecedent that does not belong to class C, and if no such antecedent exists, we resolve it to the closest preceding antecedent that belongs to class C. The reason behind the heuristic's preference for a preceding antecedent that does not belong to class C is simple: since we are resolving the mention using an oracle, we want to choose the antecedent that allows us to maximize the overall score; resolving the mention to an antecedent that does not belong to C is more likely to yield a better score than resolving it to an antecedent that belongs to C, since the former was resolved using an oracle but the latter was not. The same heuristic applies if we are trying to use the oracle to resolve a mention to a preceding cluster: we first attempt to resolve it to the closest preceding cluster containing a mention that does not belong to C, and if no such antecedent exists, we resolve it to the closest preceding cluster containing a mention that belongs to C.

| | Class | Proper nouns | | | | Common nouns | | | | Pronouns | | | | |
|---|---|---|---|---|---|---|---|---|---|---|---|---|---|---|
| | | e | p | n | na | e | p | n | na | 1+2 | G3 | U3 | oa | na |
| | % | 15.2 | 1.6 | 3.2 | 13.9 | 6.3 | 0.3 | 4.7 | 19.2 | 15.1 | 4.9 | 5.1 | 4.4 | 6.1 |
| 1 | HM | 68.3 | 33.0 | 34.4 | 63.5 | 48.1 | 55.6 | 24.7 | 68.1 | 50.7 | 46.3 | 41.7 | 23.2 | 55.7 |

Using the *Conventional* feature set

| | Class | e | p | n | na | e | p | n | na | 1+2 | G3 | U3 | oa | na |
|---|---|---|---|---|---|---|---|---|---|---|---|---|---|---|
| 2 | MP | 69.6 | 35.4 | 35.6 | 65.8 | 56.1 | 54.7 | 24.0 | 70.4 | 53.8 | 55.6 | 46.1 | 24.1 | 51.9 |
| 3 | EM | 69.9 | 35.9 | 35.9 | 65.8 | 57.3 | 55.1 | 24.3 | 70.9 | 54.2 | 56.0 | 46.4 | 24.6 | 51.7 |
| 4 | MR-P | 78.3 | 41.1 | 32.7 | 76.0 | 48.5 | 58.3 | 27.2 | 78.2 | 54.1 | 57.1 | 44.9 | 22.7 | 61.6 |
| 5 | MR-J | 79.4 | 42.5 | 33.4 | 76.4 | 48.2 | 59.0 | 27.6 | 78.5 | 54.4 | 57.7 | 45.8 | 23.0 | 62.2 |
| 6 | CR-P | 79.9 | 42.5 | 34.2 | 75.9 | 64.1 | 58.6 | 27.1 | 80.8 | 57.9 | 61.8 | 49.7 | 25.6 | 58.1 |
| 7 | CR-J | 79.9 | 43.9 | 34.4 | 76.7 | 65.0 | 59.2 | 27.4 | 82.1 | 58.6 | 62.5 | 50.7 | 26.3 | 59.7 |

Using the *Lexical* feature set

| | Class | e | p | n | na | e | p | n | na | 1+2 | G3 | U3 | oa | na |
|---|---|---|---|---|---|---|---|---|---|---|---|---|---|---|
| 8 | MP | 78.8 | 41.9 | 32.1 | 78.6 | 66.5 | 54.2 | 24.1 | 77.6 | 55.5 | 57.4 | 44.1 | 24.3 | 61.8 |
| 9 | EM | 79.1 | 41.8 | 32.4 | 78.4 | 66.5 | 54.7 | 24.1 | 77.9 | 55.7 | 57.8 | 44.1 | 24.2 | 62.1 |
| 10 | MR-P | 78.3 | 66.4 | 40.7 | 75.8 | 53.2 | 60.7 | 28.3 | 83.3 | 62.1 | 60.6 | 47.3 | 29.1 | 61.8 |
| 11 | MR-J | 79.5 | 67.1 | 41.3 | 76.3 | 54.4 | 61.1 | 28.6 | 83.5 | 62.6 | 62.3 | 47.9 | 31.3 | 62.6 |
| 12 | CR-P | 75.3 | 65.7 | 40.4 | 76.6 | 50.2 | 61.4 | 30.3 | 81.9 | 60.3 | 61.0 | 50.6 | 36.2 | 66.3 |
| 13 | CR-J | 76.4 | 68.4 | 41.1 | 77.2 | 50.8 | 63.2 | 31.1 | 83.1 | 61.9 | 62.6 | 51.8 | 37.6 | 67.0 |

Using the *Combined* feature set

| | Class | e | p | n | na | e | p | n | na | 1+2 | G3 | U3 | oa | na |
|---|---|---|---|---|---|---|---|---|---|---|---|---|---|---|
| 14 | MP | 73.8 | 40.1 | 38.8 | 67.6 | 55.9 | 54.8 | 25.0 | 73.7 | 58.6 | 60.7 | 49.3 | 27.6 | 58.1 |
| 15 | EM | 73.9 | 40.6 | 39.2 | 68.2 | 56.3 | 55.8 | 25.0 | 74.4 | 58.7 | 61.6 | 49.6 | 26.2 | 58.3 |
| 16 | MR-P | 76.4 | 50.9 | 33.3 | 80.7 | 53.4 | 55.4 | 23.2 | 78.9 | 58.3 | 56.1 | 44.3 | 25.8 | 66.4 |
| 17 | MR-J | 77.2 | 52.3 | 34.7 | 82.0 | 54.3 | 56.8 | 24.7 | 80.9 | 59.5 | 59.5 | 45.8 | 26.5 | 67.1 |
| 18 | CR-P | 78.3 | 61.3 | 41.5 | 78.3 | 60.9 | 55.4 | 24.4 | 79.3 | 62.1 | 65.5 | 51.4 | 34.7 | 62.7 |
| 19 | CR-J | 79.9 | 62.0 | 42.4 | 79.1 | 62.8 | 56.8 | 25.5 | 79.9 | 62.7 | 66.2 | 52.8 | 35.5 | 64.4 |

Table 14: B$^3$ F-measure scores of different resolution classes.

noun class (e) and all three classes that correspond to non-anaphoric mentions (na). These results indicate that it is important to take into account the non-anaphoric mentions when analyzing the performance of a coreference model. At the same time, we can see that the joint mention-ranking model does not resolve the type 'e' common nouns as well as the mention-pair model and the entity-mention model. Also, results in rows 5 and 7 indicate that the joint cluster-ranking model is better than the joint mention-ranking model due to its better handling of the type 'e' common nouns, the non-anaphoric common nouns, as well as the anaphoric pronouns.

Next, recall from Table 4 that when the *Lexical* features are used in lieu of the *Conventional* features, the mention-pair model, the entity-mention model, and the joint mention-ranking model all exhibit significant improvements in performance. For the mention-pair model and the entity-mention model, such improvements stem primarily from better handling of three proper noun classes (e,p,na), two common noun classes (e,na), and the non-anaphoric pronouns (compare rows 2 and 8 as well as rows 3 and 9 of Table 14). For the joint mention-ranking model, on the other hand, the improvements accrue from better handling of two proper noun classes (p,n), two common classes (e,na), and the anaphoric pronouns, as can be seen from rows 5 and 11. While the joint cluster-ranking model does not show overall improvement as we switch from *Conventional* to *Lexical* features (compare rows 7 and 13), the resulting models behave differently. Specifically, using the *Lexical* features, the model gets worse at handling one proper noun class (e) and one common noun class (e), but better at handling another proper noun class (n), two other common noun classes (p,n), one anaphoric pronoun class (1+2), and the non-anaphoric pronouns.

Finally, recall that when the *Combined* features are used in lieu of the *Lexical* features, all but the cluster-ranking model show a deterioration in performance. For the mention-pair model and the entity-mention model, the deterioration in performance can be attributed to poorer handling of two proper noun classes (e,na), two common noun classes (e,na), and the non-anaphoric pronouns, although they are better at handling one proper noun class (n) and the anaphoric pronouns (compare rows 8 and 14 as well as rows 9 and 15 of Table 14). Overall, poorer handling of anaphoricity appears to be a major factor responsible for the performance deterioration. For the joint mention-ranking model, the reasons for the performance deterioration are slightly different: comparing rows 11 and 17, we see its poorer handling of two proper noun classes (p,n), three common noun classes (p,n,na), and the anaphoric pronouns, although it is better at handling the non-anaphoric proper nouns and pronouns. As mentioned before, the two versions of the cluster-ranking model improve when they are trained on the *Combined* features. However, such improvements do not stem from improvements for all classes (compare rows 12 and 18 as well as rows 13 and 19). For instance, when replacing the *Lexical* features with the *Combined* features in the joint cluster-ranking model, we see improvements for two proper noun classes (e,na), one common noun class (e), and several pronoun classes (1+2,G3,U3), but performance drops for another proper noun class (p), three other common noun classes (p,n,na), and two pronoun classes (oa,na).

Overall, these results provide us with additional insights into the strengths and weaknesses of a learning-based coreference model as well as directions for future work. In particular, even if two models yield similar overall performance, they can be quite different at the resolution class level. Since there is no single coreference model that outperforms the others on all resolution classes, it may be beneficial to apply an ensemble approach, where an anaphor belonging to a particular resolution class is resolved by the model that offers the best performance for that class.

## 8. Conclusions

As Mitkov (2001, p. 122) puts it, coreference resolution is a "difficult, but not intractable problem," and researchers have been making "steady progress" on improving machine learning approaches to the problem in the past fifteen years. The progress is slow, however. Despite its deficiencies, the mention-pair model was widely thought to be the only learning-based coreference model for almost a decade. The entity-mention model and the mention-ranking model emerged only after the mention-pair model has dominated learning-based coreference research for nearly ten years. Although these two models are conceptually simple, they represent a significant departure from the mention-pair model and a new way of thinking about how alternative models of coreference can be designed. Our cluster-ranking model further advances learning-based coreference research theoretically by combining the strengths of these two models, thereby addressing two commonly cited weaknesses of the mention-pair model. It not only bridges the gap between two independent lines of learning-based coreference research — one concerning the entity-mention model and the other the mention-ranking model — that has been going on for the past few years, but also narrows the modeling gap between the sophistication of rule-based coreference models and the simplicity of learning-based coreference models. Empirically, we have shown using the ACE 2005 coreference data set that the cluster-ranking model acquired by jointly learning anaphoricity determination and coreference resolution surpasses the performance of several competing approaches, including the mention-pair model, the entity-mention model, and the mention-ranking model. Perhaps equally importantly, our cluster-ranking model is the only model considered here that can profitably exploit the information provided by two fairly different sources of information, the *Conventional* features and the *Lexical* features.

While ranking is a more natural formulation of coreference resolution than classification, ranking-based coreference models have not been more popularly used than the influential mention-pair model. One of our goals in this article is to promote the application of ranking techniques to coreference resolution. Specifically, we attempted to clarify the difference between classification-based and ranking-based coreference models by showing the constrained optimization problem that an SVM learner needs to solve for each type of models, hoping that this will help the reader appreciate the importance of ranking for coreference resolution. In addition, we have provided ample empirical evidence that ranking-based models are superior to classification-based models for coreference resolution.

Another contribution of our work lies in the empirical demonstration of the benefits of lexicalizing learning-based coreference models. While previous work showed that lexicalization only provides marginal benefits to a coreference model, we showed that lexicalization can significantly improve the mention-pair model, the entity-mention model, and the mention-ranking model, to the point where they approach or even surpass the performance of the cluster-ranking model. Interestingly, we showed that these models benefit from lexicalization the most when no conventional coreference features are used. This challenges the common belief that there is a prototypical set of linguistic features (e.g., gender and number agreement) that must be used for constructing learning-based coreference systems. In addition, our feature analysis experiments indicated that the conventional features contributed less to overall performance than the rarely studied lexical features for our joint cluster-ranking coreference model when the two types of features are used in combination.

Finally, we examined the performance of each coreference model in resolving mentions belonging to different resolution classes. We found that even if two models achieve similar overall performance, they can be quite different at the resolution class level. Overall, these results provide us with additional insights into the strengths and weaknesses of a learning-based coreference model as well as promising directions for future research.

## Bibliographic Note

Portions of this work were previously presented in a conference publication (Rahman & Ng, 2009). The current article extends this work in several ways, most notably: (1) an overview of the literature on ranking approaches to coreference resolution (Section 2); (2) a detailed explanation of the difference between classification and ranking (Section 3); (3) an investigation of the issues in lexicalizing coreference models (Section 5); and (4) an in-depth analysis of the different aspects of our coreference system (Section 7).

## Acknowledgments

The authors acknowledge the support of National Science Foundation (NSF) grant IIS-0812261. We thank the three anonymous reviewers for insightful comments and for unanimously recommending this article for publication in JAIR. Any opinions, findings, conclusions or recommendations expressed in this article are those of the authors and do not necessarily reflect the views or official policies, either expressed or implied, of NSF.